\documentclass[10pt,twocolumn,letterpaper]{article}

\usepackage{cvpr}
\usepackage{times}
\usepackage{epsfig}
\usepackage{graphicx}
\usepackage{amsmath}
\usepackage{amssymb}

\usepackage{multirow}
\usepackage{bbding}
\usepackage{pifont}
\usepackage{caption}

\usepackage{algpseudocode}
\usepackage{wrapfig}
\usepackage{tabularx, booktabs} 

\usepackage[algo2e,ruled,vlined,linesnumbered]{algorithm2e}
\usepackage{enumitem}

\usepackage{subfig}
\usepackage[toc, page]{appendix}

\usepackage[pagebackref=true,breaklinks=true,letterpaper=true,colorlinks,bookmarks=false]{hyperref}

\cvprfinalcopy 

\def\cvprPaperID{740} 

\begin{document}

\title{SCL: Towards Accurate Domain Adaptive Object Detection via \\Gradient Detach Based Stacked Complementary Losses}

 \author{Zhiqiang Shen$^1$\thanks{Equal contribution.}~, ~~Harsh Maheshwari$^{2*}$\thanks{Work done while visiting CMU.}~,~~Weichen Yao$^{1*}$,~~Marios Savvides$^1$ \\
	$^1$Carnegie Mellon University~~$^2$Indian Institute of Technology, Kharagpur\\
	\texttt{\{zhiqians,wyao2,marioss\}@andrew.cmu.edu} \\ \texttt{harshmaheshwari135@gmail.com} \\
}


\maketitle

\begin{abstract}
	Unsupervised domain adaptive object detection aims to learn a robust detector in the domain shift circumstance, where the training (source) domain is label-rich with bounding box annotations, while the testing (target) domain is label-agnostic and the feature distributions between training and testing domains are dissimilar or even totally different. In this paper, we propose a gradient detach based stacked complementary losses (SCL) method that uses detection losses as the primary objective, and cuts in several auxiliary losses in different network stages accompanying with gradient detach training to learn more discriminative representations. We argue that the prior methods~\cite{chen2018domain,he2019multi} mainly leverage more loss functions for training but ignore the interaction of different losses and also the compatible training strategy (gradient detach updating in our work). 
	Thus, our proposed method is a more syncretic adaptation learning process. We conduct comprehensive experiments on seven datasets, the results demonstrate that our method performs favorably better than the state-of-the-art methods by a significant margin. For instance, from Cityscapes to FoggyCityscapes, we achieve 37.9\% mAP, outperforming the previous art {Strong-Weak}~\cite{Saito_2019_CVPR} by 3.6\%.
\end{abstract}

\section{Introduction}

In real world scenarios, generic object detection always faces severe challenges from variations in viewpoint, background, object appearance, illumination, occlusion conditions, scene change, etc.
These unavoidable factors make object detection in domain-shift circumstance a challenging and new rising research topic in the recent years. Also, domain change is a widely-recognized, intractable problem that urgently needs to break through in reality of detection tasks, like video surveillance, autonomous driving, etc. 

\noindent{\textbf{Revisiting Domain-Shift Object Detection.}} 
Common approaches for tackling domain-shift object detection are mainly in two directions: (i) training supervised model and then fine-tuning on the target domain; or (ii) unsupervised cross-domain representation learning. The former requires additional instance-level annotations on target data, which is fairly laborious, expensive and time-consuming. So most approaches focus on the latter one but still have some challenges. The first challenge is that the representations of source and target domain data should be embedded into a common space for matching the object, such as the hidden feature space~\cite{Saito_2019_CVPR,chen2018domain}, input space~\cite{tzeng2018splat,cai2019exploring} or both of them~\cite{kim2019diversify}. The second is that a feature alignment/matching operation or mechanism for source/target domains should be further defined, such as subspace alignment~\cite{raj2015subspace}, $\mathcal{H}$-divergence and adversarial learning~\cite{chen2018domain}, MRL~\cite{kim2019diversify}, Strong-Weak alignment~\cite{Saito_2019_CVPR}, universal alignment~\cite{wang2019towards}, etc. In general, our SCL targets at these two challenges, and it is also a learning-based alignment method across domains with an end-to-end framework.

\begin{figure*}[t]
	\centering
	\subfloat[\scriptsize Non-adapted]{\includegraphics[width=0.25\textwidth, keepaspectratio]{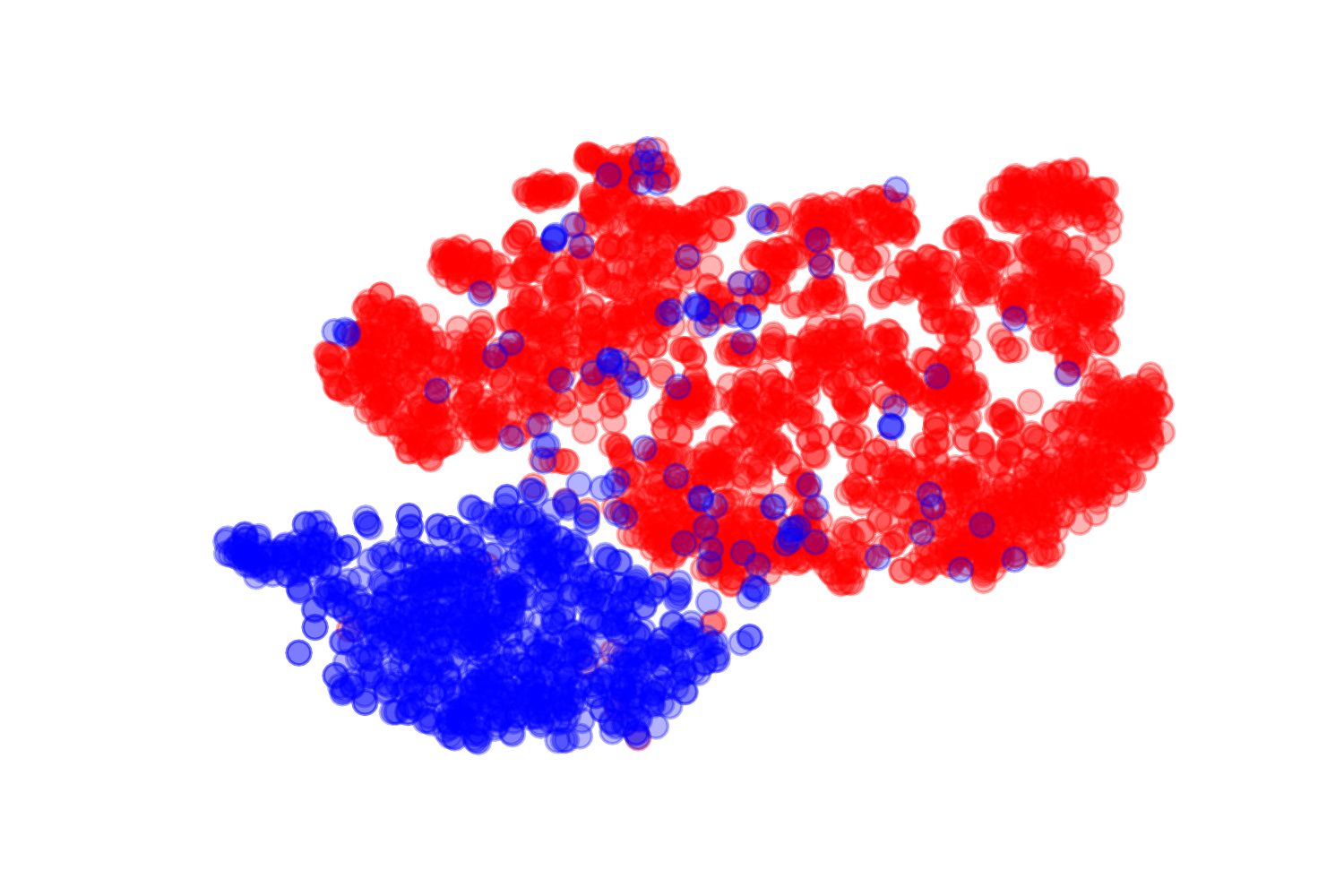}\label{fig:subfig1}}
	\subfloat[\scriptsize CVPR'18~\cite{chen2018domain}]{\includegraphics[width=0.25\textwidth, keepaspectratio]{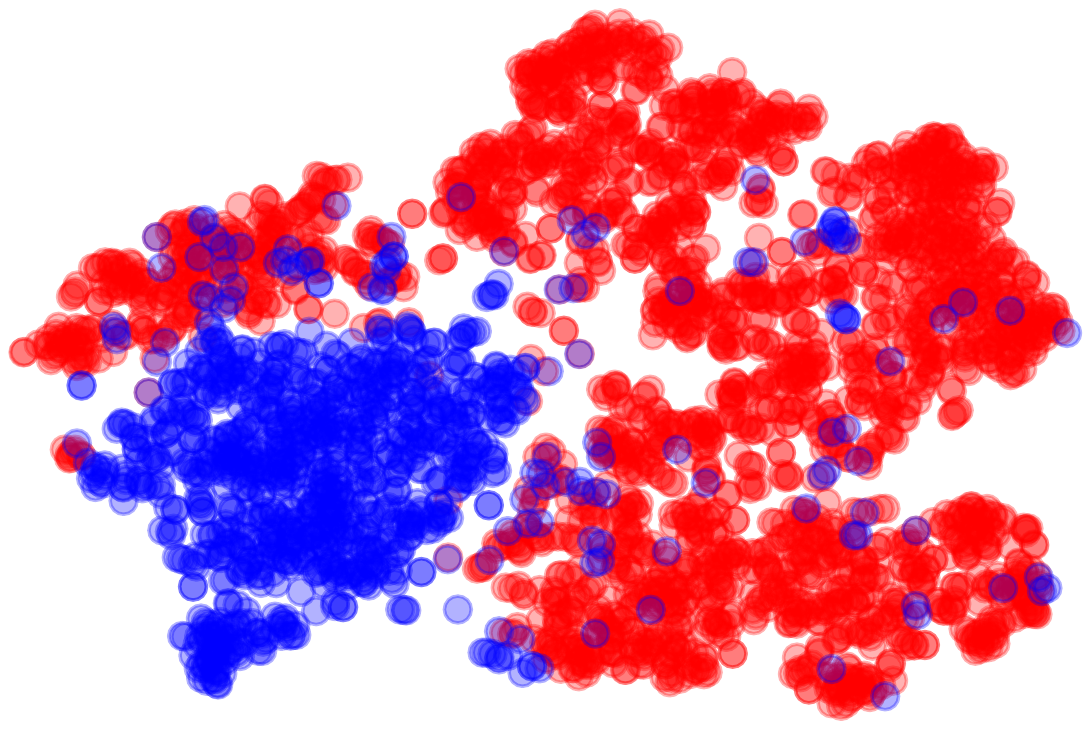}\label{fig:subfig2}}
	\subfloat[\scriptsize CVPR'19~\cite{Saito_2019_CVPR}]{\includegraphics[width=0.25\textwidth, keepaspectratio]{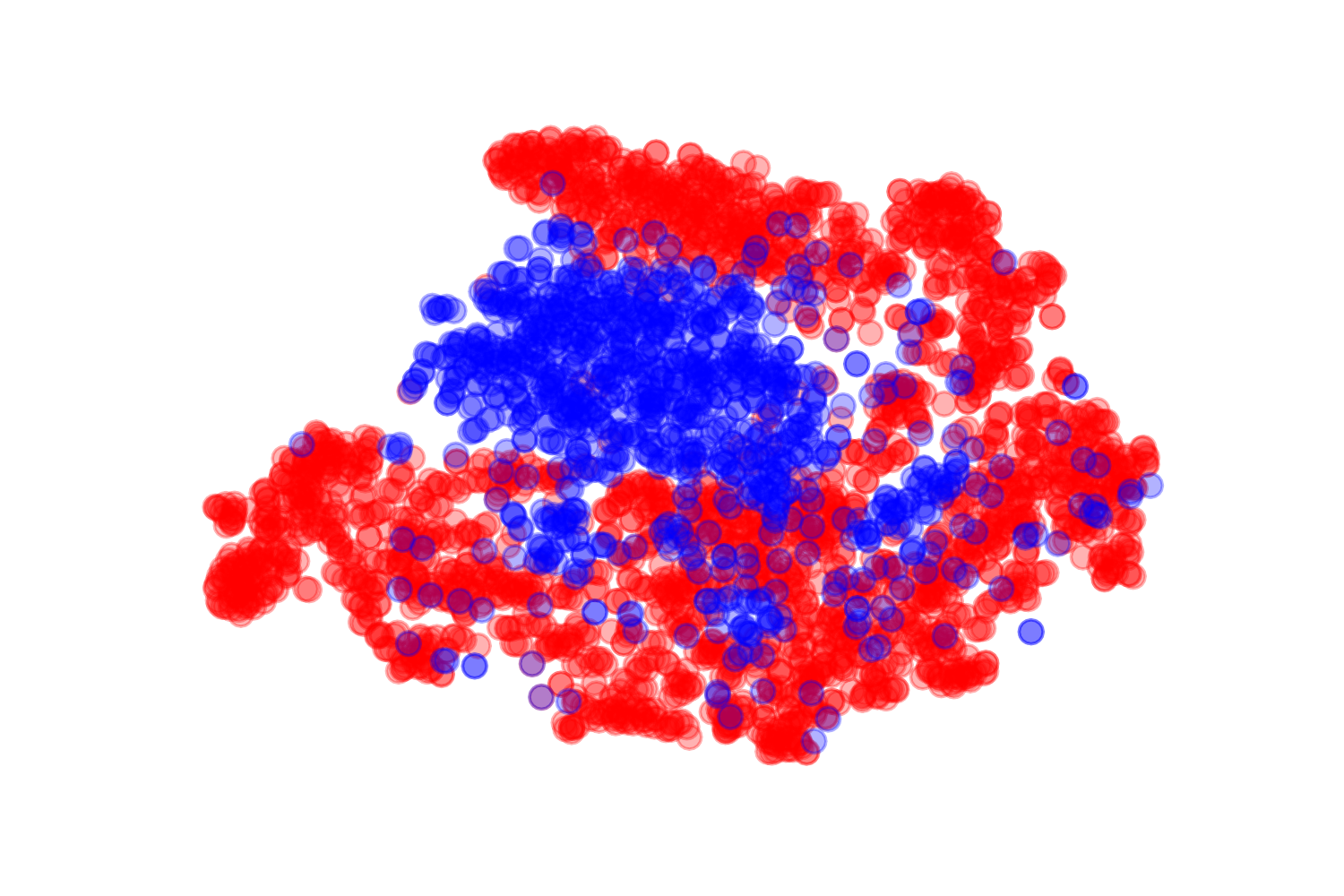}\label{fig:subfig3}}
	\subfloat[\scriptsize SCL (Ours)]{\includegraphics[width=0.25\textwidth, keepaspectratio]{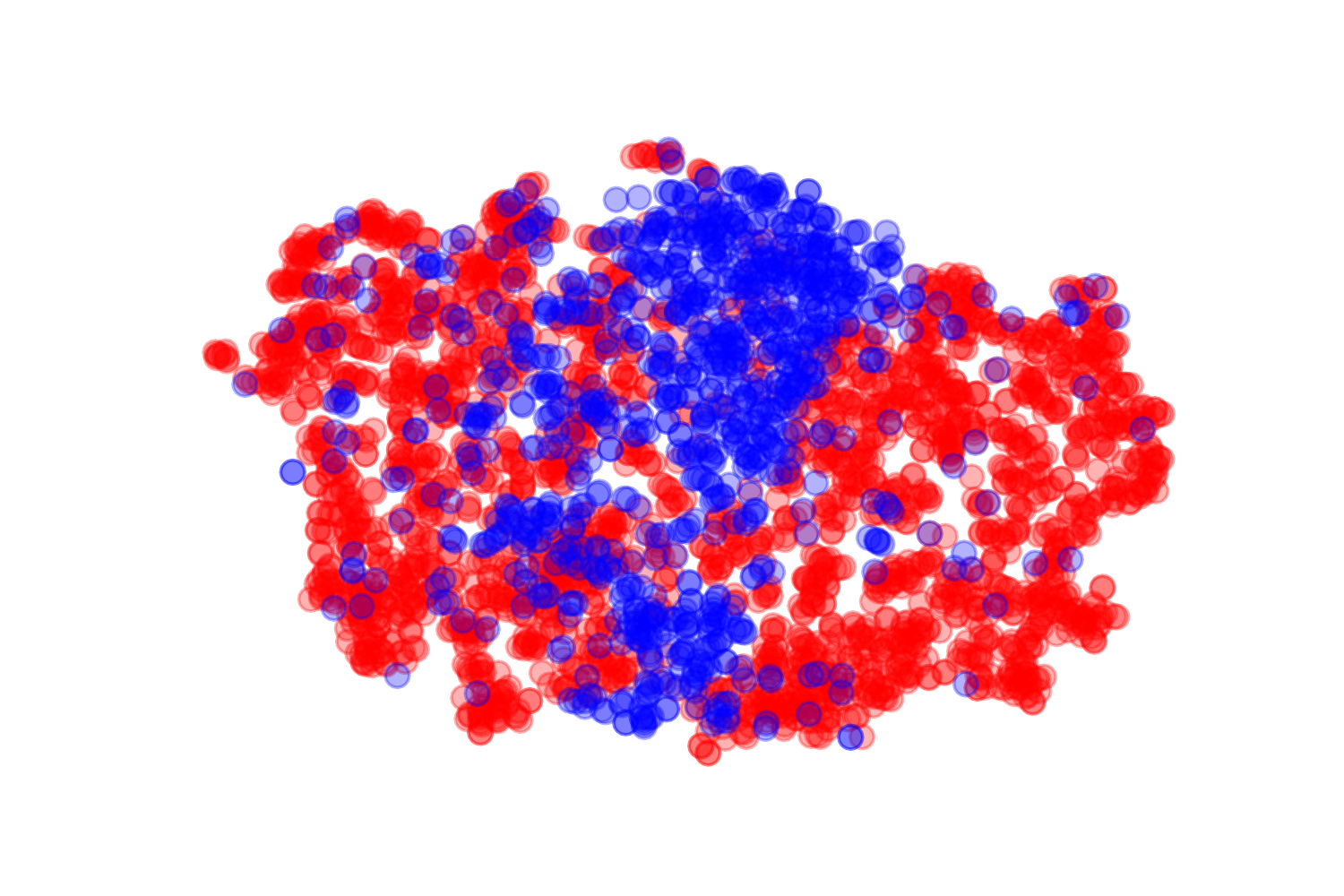}\label{fig:subfig4}}
	
	\subfloat[\scriptsize Non-adapted]{\includegraphics[width=0.25\textwidth, keepaspectratio]{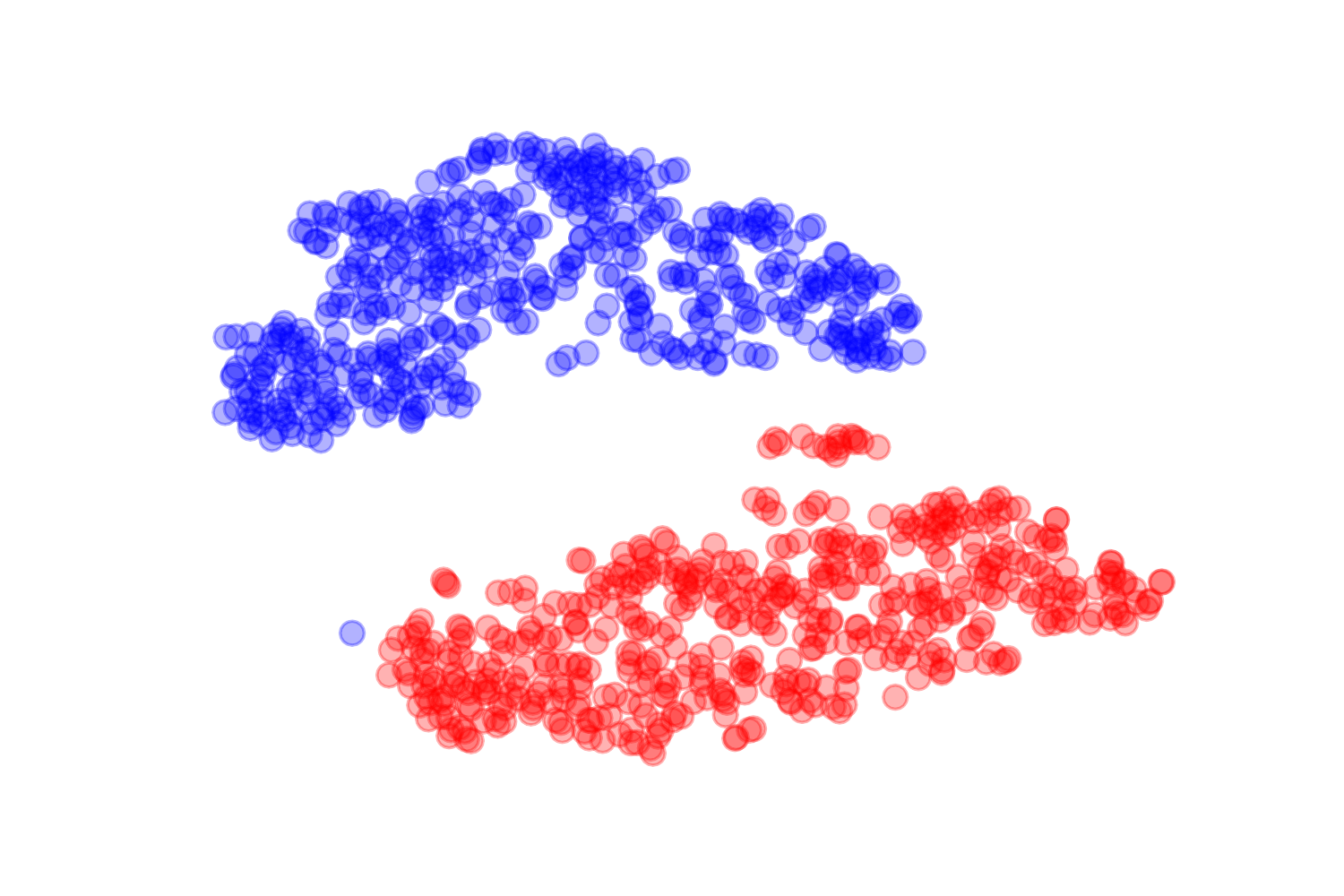}\label{fig:subfig5}}
	\subfloat[\scriptsize CVPR'18~\cite{chen2018domain}]{\includegraphics[width=0.25\textwidth, keepaspectratio]{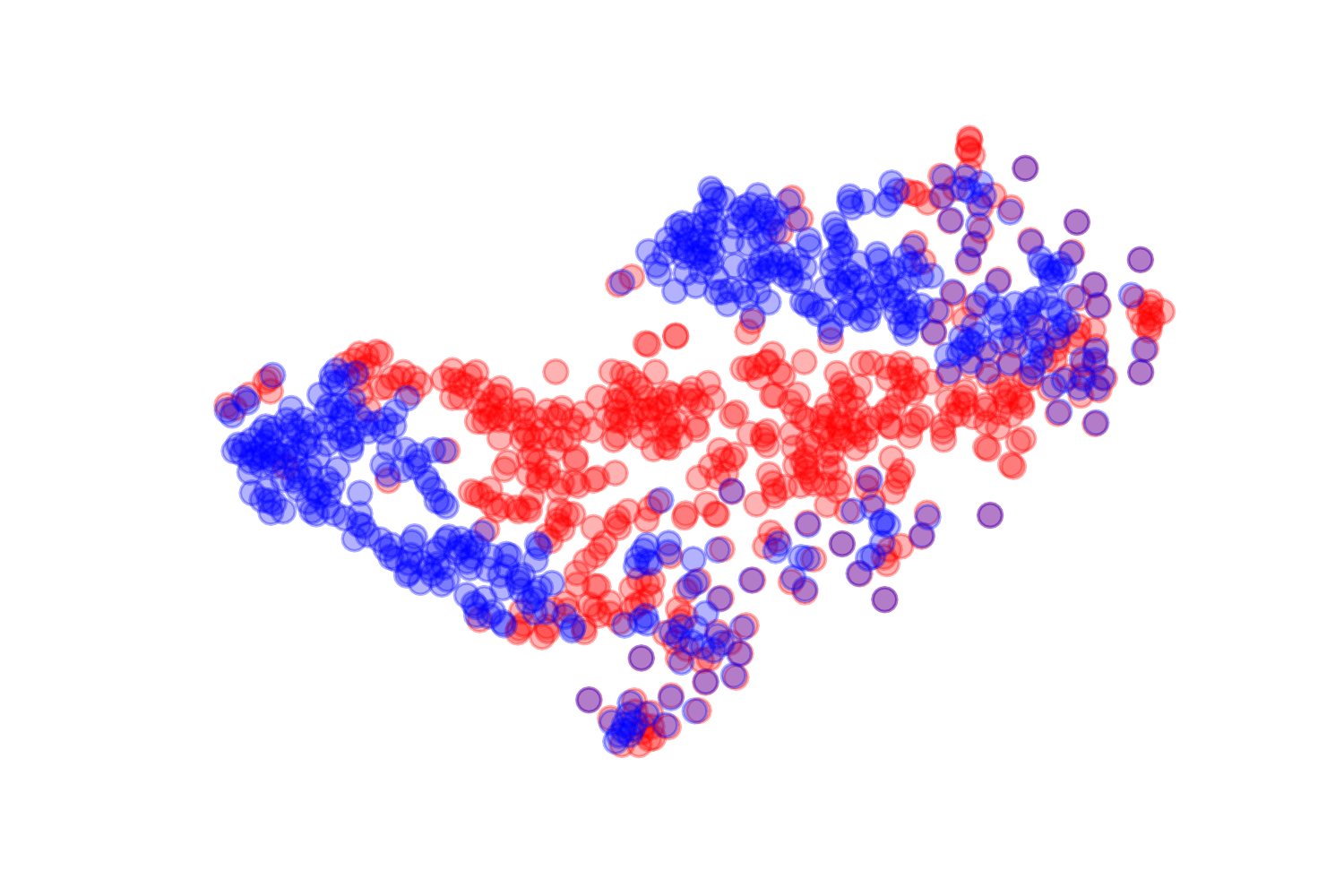}\label{fig:subfig6}}
	\subfloat[\scriptsize CVPR'19~\cite{Saito_2019_CVPR}]{\includegraphics[width=0.25\textwidth, keepaspectratio]{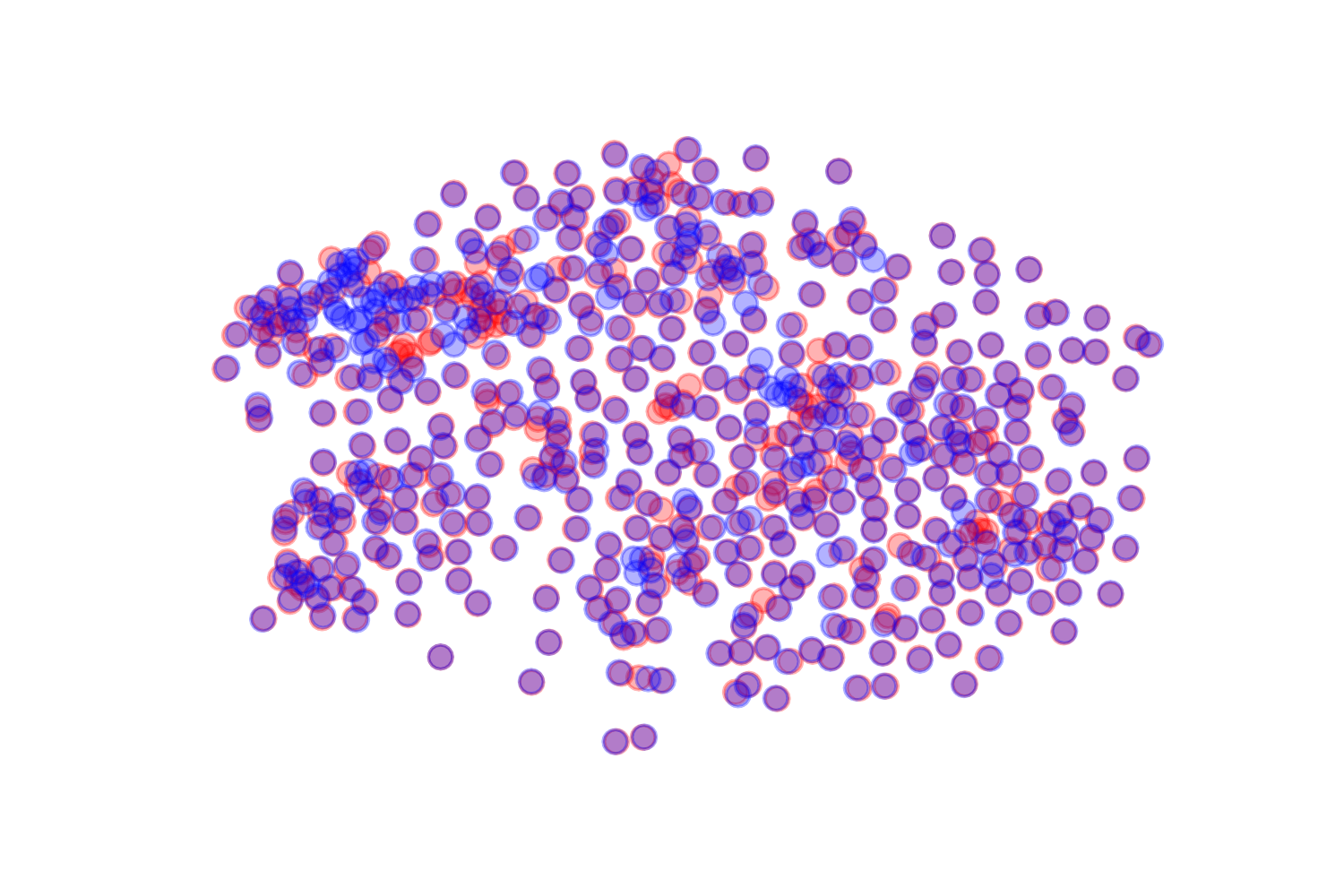}\label{fig:subfig7}}
	\subfloat[\scriptsize SCL (Ours)]{\includegraphics[width=0.25\textwidth, keepaspectratio]{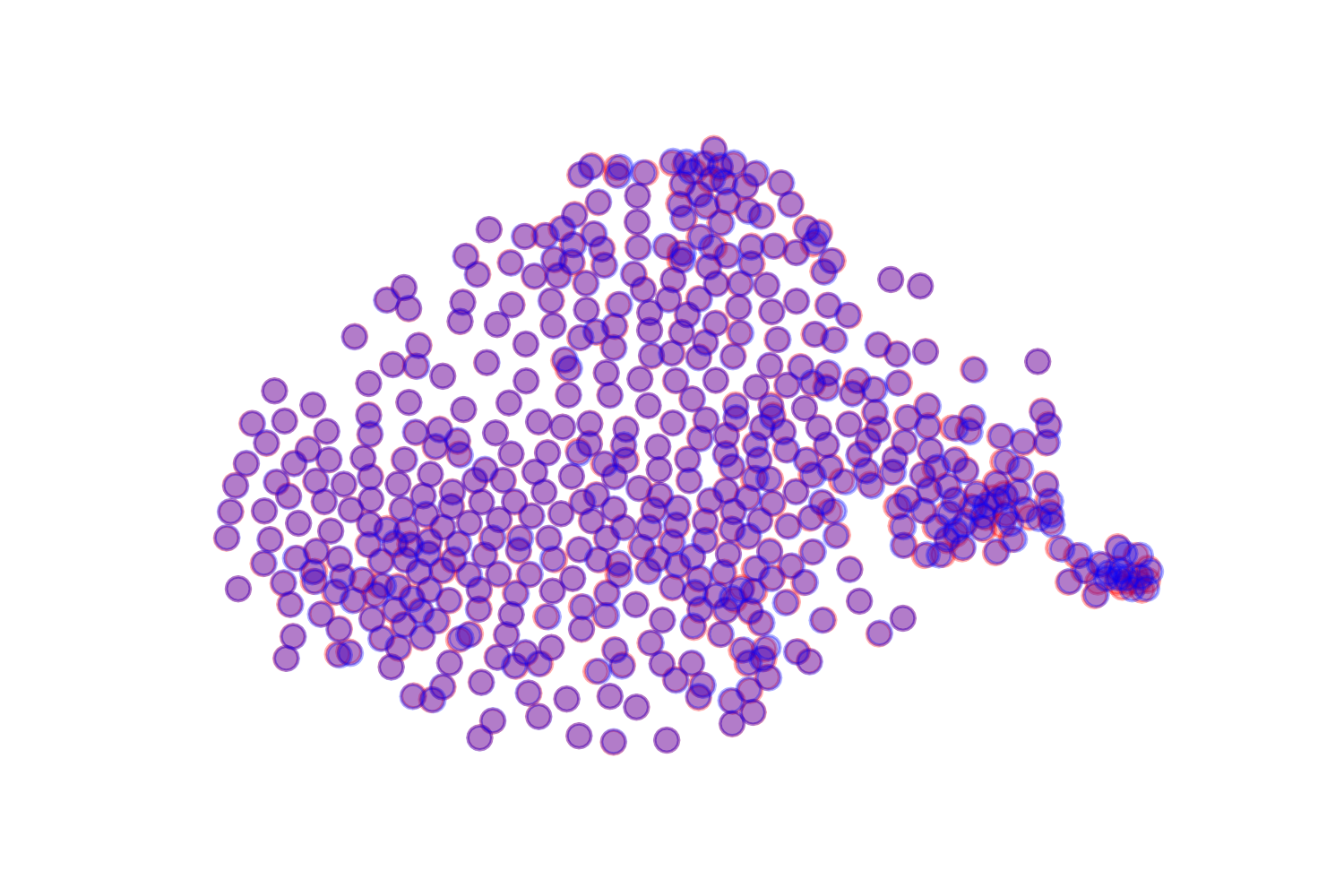}\label{fig:subfig8}}
	
	\vspace{-0.1in}
	\caption{Visualization of features from PASCAL to Clipart (first row) and from Cityscapes to FoggyCityscapes (second row) by t-SNE~\cite{maaten2008visualizing}. 
		Red indicates the source examples and blue is the target one. If source and target features locate in the same position, it is shown as light blue.
		All models are re-trained with a unified setting to ensure fair comparisons. It can be observed that our feature embedding results are consistently much better than previous approaches on either dissimilar domains (PASCAL and Clipart) or similar domains (Cityscapes and FoggyCityscapes). Best viewed in color and zoom in.}
	\label{visualization}
	\vspace{-0.05in}
\end{figure*}

\noindent{\textbf{Our Key Ideas.}} The goal of this paper is to introduce a simple design that is specific to convolutional neural network optimization and improves its training on tasks that adapt on discrepant domains. Unsupervised domain adaptation for recognition has been widely studied by a large body of previous literature~\cite{ganin2016domain,long2016unsupervised,tzeng2017adversarial,panareda2017open,hoffman2018cycada,murez2018image,zhao2019learning,wu2019domain}, our method more or less draws merits from them, like aligning source and target distributions with adversarial learning (domain-invariant alignment). However, object detection is a technically different problem from classification, since we would like to focus more on the object of interests (regions). 


Some recent work~\cite{zhu2019adapting} has proposed to conduct alignment only on local regions so that to improve the efficiency of model learning. While this operation may cause a deficiency of critical information from context. Inspired by strong-weak/multi-feature alignment~\cite{Saito_2019_CVPR,zhang2018collaborative,he2019multi} which proposed to align corresponding local-region on shallow layers with a small respective field (RF) and align image-level features on deep layers with large RF, we extend this idea by studying diverse complementary objectives and their potential assembly for domain adaptive circumstance. We observe that domain adaptive object detection is supported dramatically by the deep supervision, however, the diverse supervisions should be applied in a controlled manner, including the cut-in locations, loss types, orders, updating strategy, etc., which is one of the contributions of this paper.
Furthermore, our experiments show that even with the existing objectives, after elaborating the different combinations and training strategy, our method can obtain competitive results. 
By plugging-in a new sub-network that learns the context features independently with gradient detach updating strategy in a hierarchical manner, we obtain the best results on several domain adaptive object detection benchmarks.

\noindent{\textbf{The Relation to Complement Objective Training~\cite{chen2018complement} and Deep Supervision~\cite{lee2015deeply}.}} {COL}~\cite{chen2018complement} proposed to involve additional loss function that complements the primary objective 
which is moderately similar to our goal in spirit. The cross entropy in COL is used as the primary objective $\mathbf{H_p}$:
\begin{equation}
\begin{aligned} \mathbf{H_p}(\mathbf{y}, \hat{\mathbf{y}}) &=-\frac{1}{N} \sum_{i=1}^{N} \mathbf{y}_{i}^{T} \cdot \log \left(\hat{\mathbf{y}}_{i}\right)  \end{aligned}
\end{equation}
where ${\mathbf{y}_i} \in {\{ 0,1\} ^D}$ is the label of the $i$-th sample in one-hot representation and ${\hat {\mathbf{y}}_i} \in {[0,1]^D}$ is the predicted probabilities.

Th complement entropy $\mathbf{H_c}$ is defined in COT~\cite{chen2018complement} as the average of sample-wise entropies over complement classes in a mini-batch:
\begin{equation}
\begin{aligned} \mathbf{H_c}\left(\hat{\mathbf{y}}_{\overline{c}}\right) &=\frac{1}{N} \sum_{i=1}^{N} \mathcal{H}\left(\hat{\mathbf{y}}_{{i} \overline{c}}\right) \end{aligned}
\end{equation}
where $\mathcal H$ is the entropy function. $\hat{\mathbf{y}}_{\overline{c}}$ is the predicted probabilities of complement classes $\overline{c}$. The training process is that: for each iteration of training, 1) update parameters by $\mathbf{H_p}$ first; then 2) update parameters by $\mathbf{H_c}$. Different from COL, we don't use the alternate strategy but update the parameters simultaneously using gradient detach strategy with primary and complement objectives. Since we aim to let the network enable to adapt on both source and target domain data and meanwhile still being able to distinguish objects from them. Thus our complement objective design is quite different from COT. We will describe with details in Section~\ref{method}.

In essence, our method is more likely to be the deeply supervised formulation~\cite{lee2015deeply} that backpropagation of error now proceeds not only from the final layer but also simultaneously from our intermediate complementary outputs. While DSN is basically proposed to alleviate ``vanishing'' gradient problem, here we focus on how to adopt these auxiliary losses to promote to mix two different domains through domain classifiers for detection. Interestingly, we observe that diverse objectives can lead to better generalization for network adaptation. Motivated by this, we propose {\bf S}tacked {\bf C}omplementary {\bf L}osses (SCL), a simple yet effective approach for domain-shift object detection. Our SCL is fairly easy and straightforward to implement, but can achieve remarkable performance. We conjecture that previous approaches that focus on conducting domain alignment on high-level layers only~\cite{chen2018domain} cannot fully adapt shallow layer parameters to both source and target domains (even local alignment is applied~\cite{Saito_2019_CVPR}) which restricts the ability of model learning. Also, gradient detach is a critical part of learning with our complementary losses. We further visualize the features obtained by non-adapted model, DA~\cite{chen2018domain}, Strong-Weak~\cite{Saito_2019_CVPR} and ours, features are from the last layer of backbone before feeding into the Region Proposal Network (RPN). As shown in Figure~\ref{visualization}, it is obvious that the target features obtained by our model are more compactly matched with the source domain than any other models.


\noindent{\textbf{Contributions.}} Our contributions are three-fold.
\begin{itemize}[leftmargin=5.5mm]
	\vspace{-0.05in}
	\addtolength{\itemsep}{-0.05in}
	\item We study {\bf the interaction of multi-loss (deep supervision with complement objective) and gradient detach (training strategy for maximizing context information)} in an end-to-end learnable framework for the challenging domain adaptive object detection task.
	\item  We provide a step-by-step ablation study to empirically verify the effectiveness of each component and design in our framework. Thus, this work gives intuitive and practical guidance for building a high performance framework with multi-objective learning on domain adaptive object detection.
	\item To the best of our knowledge, this is a pioneer work to investigate the influence of diverse loss functions and gradient detach for domain adaptive object detection. Our method achieves the highest accuracy on several domain adaptive or cross-domain object detection benchmarks\footnote{Our code and models are available at: \url{https://github.com/harsh-99/SCL}.}.
\end{itemize}


\begin{figure*}[t]
	\centering
	\hspace{1.1cm}
	\includegraphics[width=0.89\textwidth]{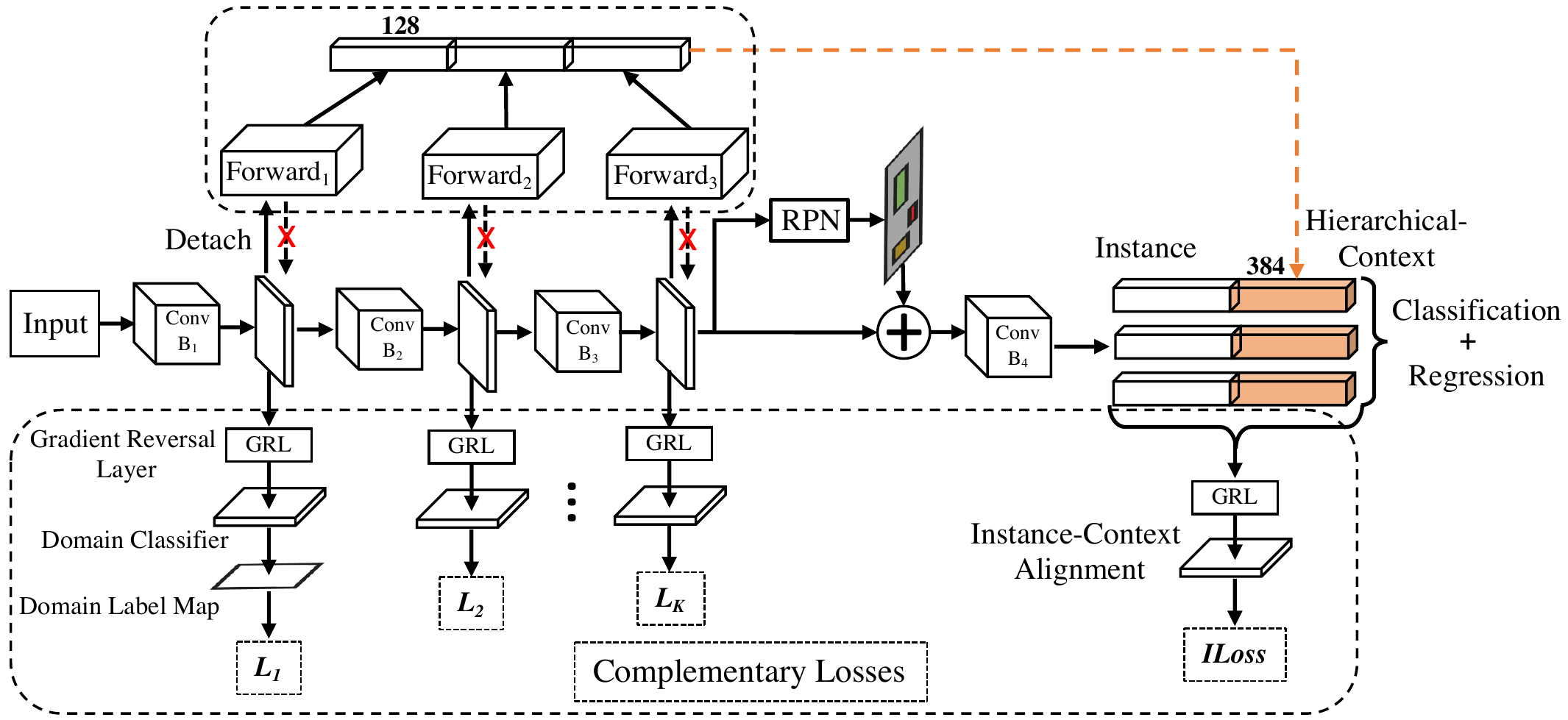}
	\vspace{-0.1in}
	\caption{Overview of our SCL framework. More details please refer to Section~\ref{method}.}
	\label{framework}
\end{figure*}

\section{Methodology} \label{method} 

Following the common formulation of domain adaptive object detection, we define a {\em source domain} $\mathcal{S}$ where annotated bound-box is available, and a {\em target domain $\mathcal{T}$} where only  the image can be used in training process without any labels. Our purpose is to train a robust detector that can adapt well to both source and target domain data, i.e., we aim to learn a  {\em domain-invariant} feature representation that works well for detection across two different domains.
\subsection{Multi-Complement Objective Learning}
As shown in Figure~\ref{framework}, we focus on the complement objective  learning and let $\mathcal{S}=\{(\mathbf{x}_i^{(s)}, \mathbf{y}_i^{(s)})\}$ where $\mathbf{x}_i^{(s)} \in \mathcal{R}^n$ denotes an image, $\mathbf{y}^{(s)}_i$ is the corresponding bounding box and category labels for sample $\mathbf{x}^{(s)}_i$, and $i$ is an index. Each label $\mathbf{y}^{(s)}=(y_\mathbf{c}^{(s)},y_\mathbf{b}^{(s)})$ denotes a class label $y_\mathbf{c}^{(s)}$ where $\mathbf{c}$ is the category, and a 4-dimension bounding-box coordinate $y_\mathbf{b}^{(s)} \in \mathcal{R}^4$. For the target domain we only use image data for training, so $\mathcal{T}=\{\mathbf{x}_i^{(t)}\}$. We define a recursive function for layers $\mathbf{k}=1,2,\dots,\mathbf{K}$ where we cut in complementary losses:
\begin{equation}
\begin{array}{l}{\hat \Theta_{\mathbf{k}}=\mathcal{F}\left(\mathbf{Z}_{\mathbf{k}}\right), \text { and } \mathbf{Z}_{0} \equiv \mathbf{x}} \end{array}
\end{equation}
where $\hat \Theta_{\mathbf{k}}$ is the feature map produced at layer $\mathbf{k}$, $\mathcal{F}$ is the function to generate features at layer $\mathbf{k}$ and $\mathbf{Z}_{\mathbf{k}}$ is input at layer $\mathbf{k}$. We formulate the complement loss of domain classifier $\mathbf{k}$ as follows:
\begin{equation}\label{k}
\begin{gathered} \mathcal{L}_{\mathbf{k}}\left(\hat{\Theta}^{(s)}_{\mathbf{k}}, \hat{\Theta}^{(t)}_{\mathbf{k}} ; \mathbf{D}_{\mathbf{k}}\right)={\cal L}_\mathbf{k}^{(s)}({{\hat \Theta }^{(s)}_{\mathbf{k}}};{\mathbf{D}_{\bf{k}}}) + {\cal L}_\mathbf{k}^{(t)}({{\hat \Theta }^{(t)}_{\mathbf{k}}};{\mathbf{D}_{\bf{k}}}) \\ = \mathbb{E}\left[\log \left(\mathbf{D}_{\mathbf{k}}\left(\hat{\Theta}^{(s)}_{\mathbf{k}}\right)\right)\right]  + \mathbb{E}\left[\log \left(1-\mathbf{D}_{\mathbf{k}}\left(\hat{\Theta}^{(t)}_{\mathbf{k}}\right)\right)\right] \end{gathered}
\end{equation}
where $\mathbf{D}_\mathbf{k}$ is the $\mathbf{k}$-th domain classifier or discriminator. $\hat{\Theta}^{(s)}_{\mathbf{k}}$ and $\hat{\Theta}^{(t)}_{\mathbf{k}}$ denote feature maps from source and target domains respectively. Following~\cite{chen2018domain,Saito_2019_CVPR}, we also adopt gradient reverse layer (GRL)~\cite{ganin2015unsupervised} to enable adversarial training where a GRL layer is placed between the domain classifier and the detection backbone network. During backpropagation, GRL will reverse the gradient that passes through from domain classifier to detection network. 

For our instance-context alignment loss ${\mathcal{L}_{{\mathbf{ILoss}}}}$, we take the instance-level representation and context vector as inputs. The instance-level vectors are from RoI layer that each vector focuses on the representation of local object only. The context vector is from our proposed sub-network that combine hierarchical global features. We concatenate instance features with same context vector. Since context information is fairly different from objects, joint training detection and context networks will mix the critical information from each part, here we proposed a better solution that uses detach strategy to update the gradients.  We will introduce it with details in the next section. Aligning instance and context representation simultaneously can help to alleviate the  variances of object appearance, part deformation, object size, etc. in instance vector and illumination, scene, etc. in context vector. We define $d_i$ as the domain label of $i$-th training image where $d_i=1$ for the source and $d_i=0$ for the target, so the instance-context alignment loss can be further formulated as:
\begin{equation}
\begin{aligned} {\mathcal{L}_{{\mathbf{ILoss}}}} =  - \frac{1}{{{N_s}}}\sum\limits_{i = 1}^{{N_s}} {\sum\limits_{i,j} {(1 - d_i)} \log {{\mathbf{P}}_{(i,j)}}}  \\ \quad- \frac{1}{{{N_t}}}\sum\limits_{i = 1}^{{N_t}} {\sum\limits_{i,j} {d_i\log \left( {1 - {{\mathbf{P}}_{(i,j)}}} \right)} } \end{aligned}
\end{equation}
where $N_s$ and $N_t$ denote the numbers of source and target examples. $\mathbf{P}_{(i,j)}$ is the output probabilities of the instance-context domain classifier for the $j$-th region proposal in the $i$-th image. So our total {\bf SCL} objective $\mathcal{L}_{\mathbf{SCL}}$ can be written as:
\begin{equation}
{\mathcal{L}_{\mathbf{SCL}}} = \sum\limits_{\mathbf{k} = 1}^\mathbf{K} {{\mathcal{L}_{\mathbf{k}}}}  + {\mathcal{L}_{\mathbf{ILoss}}}
\end{equation}


\subsection{Gradients Detach Updating}

In this section, we introduce a simple yet effective detach strategy which prevents the flow of gradients from context sub-network through the detection backbone path. We find this can help to obtain more discriminative context and we show empirical evidence (see Figure~\ref{heatmaps}) that this path carries information with diversity and hence gradients from this path getting suppressed is superior for such task.

As aforementioned, we define a sub-network to generate the context information from early layers of detection backbone. Intuitively, instance and context will focus on perceptually different parts of an image, so the representations from either of them should also be discrepant. However, if we train with the conventional process, the companion sub-network will be updated jointly with the detection backbone, which may lead to an indistinguishable behavior from these two parts. To this end, in this paper we propose to suppress gradients during backpropagation and force the representation of context sub-network to be dissimilar to the detection network, as shown in Algorithm~\ref{alg:detach}.
To our best knowledge, this may be the first work to show the effectiveness of gradient detach that can help to learn better context representation for domain adaptive object detection. 
The details of our context sub-network architecture are illustrated in Appendix~\ref{sec:sfp}.

\begin{algorithm2e}[h]
	\caption{Backward Pass of Our Detach Algorithm}
	\label{alg:detach}
	{\bf INPUT:} $\mathbf{G}_{\bf c}$ is gradient of context network, $\mathbf{G}_{\bf d}$ is the gradient of detection network, $\mathcal{L}_{det}$ is the detection objective, $\mathcal{L}_\mathbf{SCL}$ is the complementary objective; 
	
	\For{$t \gets 1$ \textbf{to} $n_{train\_steps}$} {
		{1. Update context net by detection and instance-context objectives:~$\mathcal{L}_{det}$(w/o $\mathcal{L}_{rpn}$)+$\mathcal{L}_\mathbf{ILoss}$}
		
		{2. $\mathbf{G}_{\bf d} \gets$ stop-gradient($\mathbf{G}_{\bf c}$;$\mathcal{L}_{det}$)}
		
		{3. Update detection net by detection and complementary objectives:~$\mathcal{L}_{det}$+$\mathcal{L}_\mathbf{SCL}$}
	}
\end{algorithm2e}

\subsection{Framework Overall}

Our detection part is based on the Faster RCNN~\cite{ren2015faster}, including the Region Proposal Network (RPN) and other modules. This is a conventional practice in many adaptive detection works. The objective of the detection loss is summarized as:
\begin{equation}
{\mathcal{L}_{det}} = {\mathcal{L}_{rpn}} + {\mathcal{L}_{cls}} + {\mathcal{L}_{reg}}
\end{equation}
where ${\mathcal{L}_{cls}}$ is the classification loss and ${\mathcal{L}_{reg}}$  is the bounding-box regression loss.
To train the whole model using SGD, the overall objective function in the model is:
\begin{equation} \label{lambda}
\min _{\mathcal{F}, \mathbf{R}} \max _{\mathbf{D}} \mathcal{L}_{det}(\mathcal{F}(\mathbf{Z}), \mathbf{R})-\lambda \mathcal{L}_\mathbf{SCL}(\mathcal{F}(\mathbf{Z}), \mathbf{D})
\end{equation}
where $\lambda$ is the trade-off coefficient between detection loss and our complementary loss. $\mathbf{R}$ denotes the RPN and other modules in Faster RCNN. 

\section{Empirical Results}

\noindent{\textbf{Datasets.}}
We evaluate our approach in three different domain shift scenarios: (1) Similar Domains; (2) Discrepant Domains; and (3) From Synthetic to Real Images.
All experiments are conducted on seven domain shift datasets: Cityscapes~\cite{cordts2016cityscapes} to FoggyCityscapes~\cite{sakaridis2018semantic}, Cityscapes to KITTI~\cite{Geiger2012CVPR}, KITTI to Cityscapes, INIT Dataset~\cite{shen2019towards}, PASCAL~\cite{everingham2010pascal} to Clipart~\cite{inoue2018cross}, PASCAL to Watercolor~\cite{inoue2018cross}, GTA (Sim 10K)~\cite{johnson2016driving} to Cityscapes.

\noindent{\textbf{Implementation Details.}} In all experiments, we resize the shorter side of the image to 600 following~\cite{ren2015faster,Saito_2019_CVPR} with ROI-align~\cite{he2017mask}. We train the model with SGD optimizer and the initial learning rate is set to $10^{-3}$, then divided by 10 after every 50,000 iterations.
Unless otherwise stated, we set $\lambda$ as 1.0 and $\gamma$ as 5.0, and we use $\mathbf{K}=3$ in our experiments (the analysis of hyper-parameter $\mathbf{K}$ is shown in Table~\ref{tab:ablation_k}). We report mean average precision (mAP) with an IoU threshold of 0.5 for evaluation. Following~\cite{chen2018domain,Saito_2019_CVPR}, we feed one labeled source image and one unlabeled target one in each mini-batch during training. SCL is implemented on PyTorch platform, we will release our code and models.

\renewcommand{\arraystretch}{1.03}
\setlength{\tabcolsep}{.2em}
\begin{table*}[t]
	\caption{Ablation study (\%) on Cityscapes to FoggyCityscapes (we use 150m visibility, the densest one) adaptation. Please refer to Section~\ref{ablation} for more details.}
	\label{ablation_foggy}
	\vspace{-1.8ex}
	\centering
	\resizebox{0.95\textwidth}{!}{%
		\begin{tabular}{l|c|ccc|c|c|cccccccc|c}
			\toprule[1.5pt]
			\multirow{2}{*}{} & \multirow{2}{*}{} & \multicolumn{1}{r}{\multirow{2}{*}{}} & \multicolumn{1}{r}{\multirow{2}{*}{}} & \multicolumn{1}{r|}{\multirow{2}{*}{}} & \multirow{2}{*}{} & \multirow{2}{*}{}& \multicolumn{9}{c}{AP on a target domain}                                    \\
			Method     &       Context     & \multicolumn{1}{r}{$L_1$}                    & \multicolumn{1}{r}{$L_2$}                    & \multicolumn{1}{r|}{$L_3$}  &    ILoss     &  Detach    & person & rider & car   & truck & bus   & train & mcycle & bicycle & \bf mAP  \\ \hline
			Faster RCNN (Non-adapted)&&&&&&&24.1&33.1&34.3&4.1&22.3&3.0&15.3&26.5&20.3\\ 
			DA (CVPR'18)&$\checkmark$&&&&&&25.0&31.0&40.5&22.1&35.3&20.2&20.0&27.1& 27.6 \\ 
			MAF~\cite{he2019multi} (ICCV'19)&&&&&& &28.2& 39.5&43.9&23.8& 39.9& 33.3& 29.2& 33.9&34.0\\ 
			Strong-Weak (CVPR'19)&$\checkmark$&&&&& &29.9&42.3&43.5&24.5&36.2&32.6&30.0& 35.3&  34.3\\  
			Diversify\&match~\cite{kim2019diversify} (CVPR'19)&&&&&& &30.8& 40.5& 44.3& 27.2& 38.4& 34.5& 28.4& 32.2&34.6\\ \hline 
			Strong-Weak (Our impl. w/ VGG16)&$\checkmark$&&&&& &30.0&40.0&43.4&23.2&40.1&34.6&27.8& 33.4&34.1\\  
			Strong-Weak (Our impl. w/ Res101)&$\checkmark$&&&&&&29.1&41.2&43.8&26.0&43.2&27.0&26.2& 30.6&33.4\\ \hline 
			&\ding{55}&$LS$&$FL$&\ding{55}& \ding{55}&\ding{55}&29.6&42.2&43.4&23.1&36.4&31.5&25.1&30.5&32.7 \\
			&\ding{55}&$LS$&$CE$&$FL$& \ding{55}&\ding{55}&29.0&41.4&43.9&24.6&46.5&28.5&27.0&32.8&34.2\\
			&\ding{55}&$LS$&$CE$&$FL$&$FL$&\ding{55}&28.6&44.0&44.2&25.2&42.9&31.1&27.4&33.0&34.5\\ \hline
			&$\checkmark$&$LS$&$FL$&\ding{55}& \ding{55}&\ding{55}&28.5&42.6&43.8&23.2&41.6&24.9&28.3&30.3&32.9 \\
			&$\checkmark$&$LS$&$FL$&$FL$& \ding{55}&\ding{55}&28.6&41.8&43.8&27.9&43.3&24.0&28.7&31.3&33.7\\
			&$\checkmark$&$LS$&$LS$&$FL$& \ding{55}&\ding{55}&28.8&\bf 45.5&44.3&28.6&44.6&29.1&27.8&31.4&35.0 \\
			&$\checkmark$&$LS$&$CE$&$FL$& \ding{55}&\ding{55}&29.6&42.6&42.6&28.4&46.3&31.0&28.4&33.0&35.3\\ 
			&$\checkmark$&$LS$&$CE$&$FL$& \ding{55}&$\checkmark$&30.0&42.7&44.2&30.0&\bf 50.2&34.1&27.1&32.2&36.3\\ \cline{2-16} 
			
			&$\checkmark$&$LS$&$FL$&$FL$& $FL$& \ding{55}&26.3&42.8&44.2&26.7&41.6&36.4& 29.2&30.9&34.8\\
			&$\checkmark$&$LS$&$LS$&$FL$& $FL$&$\checkmark$&29.5&43.2&44.2&27.0&42.1&33.3&29.4&30.6&34.9\\ 
			&$\checkmark$&$LS$&$FL$&$FL$& $FL$&$\checkmark$&29.7&43.6&43.7&26.6&43.8&33.1& 30.7&31.5&35.3
			\\ \cline{2-16} 
			&$\checkmark$&$LS$&$CE$&$FL$& $CE$&\ding{55}&28.3&41.9&43.1&25.4&45.1&35.5&26.7&31.6&34.7\\
			&$\checkmark$&$LS$&$CE$&$FL$& $FL$&\ding{55}&29.8&43.9&44.0&29.4&46.3&30.0&31.8&31.8&35.8
			\\ 
			&$\checkmark$&$LS$&$CE$&$FL$& $CE$&$\checkmark$ &29.0&42.5&43.9&28.9&45.7&42.4&26.4&30.5&36.2
			\\ 
			&$\checkmark$& $LS$& $CE$& $FL$&$FL$     &  $\checkmark$  &  30.7 & 44.1 & 44.3 & 30.0 & 47.9 &\bf 42.9 & 29.6 & 33.7 & \bf 37.9   \\ \hline
			Our full model w/ VGG16&$\checkmark$&$LS$&$CE$&$FL$& $FL$&$\checkmark$ &\bf 31.6&44.0&\bf 44.8&\bf 30.4&41.8&40.7&\bf 33.6&\bf 36.2&\bf 37.9\\ \hline
			Upper Bound~\cite{Saito_2019_CVPR}                      &         --                & --                                       & --                                      & --                                       &         --     &  --   & 33.2 & 45.9 & 49.7 & 35.6 & 50.0 & 37.4 & 34.7 & 36.2&  40.3  \\ \bottomrule[1.5pt]
	\end{tabular}}
	\\
	{ \em LS: Least-squares Loss; CE: Cross-entropy Loss; FL: Focal Loss; ILoss: Instance-Context Alignment Loss.}
	\vspace{-.1in}
\end{table*}

\subsection{How to Choose Complementary Losses in Our Framework}
There are few pioneer works for exploring the combination of different losses for domain adaptive object detection, hence we conduct extensive ablation study for this part to explore the intuition and best collocation of our SCL method. We mainly adopt three losses which have been introduced in many literature. Since they are not our contribution so here we just give some brief formulations below.

\noindent{\textbf{Cross-entropy (CE) Loss.}} CE loss measures the performance of a classification model whose output is a probability value. It increases as the predicted probability diverges from the actual label:
\begin{equation}
\mathcal{L}_\mathbf{CE}(p_\mathbf{c})=- \sum\limits_{\mathbf{c} = 1}^\mathbf{C} {{y_\mathbf{c}}} \log \,{p_\mathbf{c}}
\end{equation}
where $p_\mathbf{c} \in [0,1]$ is the predicted probability observation of $\mathbf{c}$ class. $y_\mathbf{c}$ is the $\mathbf{c}$ class label.

\noindent{\textbf{Weighted Least-squares (LS) Loss.}} Following~\cite{Saito_2019_CVPR}, we adopt LS loss to stabilize the training of the domain classifier for aligning low-level features. The loss is designed to align each receptive field of features with the other domain.
The least-squares loss is formulated as:
\begin{equation}
\begin{aligned}
{\mathcal{L}_\mathbf{LS}} = \alpha{\mathcal{L}^{(s)}_{loc}} + \beta{\mathcal{L}^{(t)}_{loc}}  
= \frac{\alpha}{{HW}} {\sum\limits_{w = 1}^W {\sum\limits_{h = 1}^H \mathbf D } } \left( {{{\hat \Theta }^{(s)}}} \right)_{wh}^2 \hfill 
\\ + \frac{\beta}{{HW}} {\sum\limits_{w = 1}^W {\sum\limits_{h = 1}^H {{{\left( {1 - \mathbf{D}{{\left( {{{\hat \Theta }^{(t)}}} \right)}_{wh}}} \right)}^2}} } }  \hfill \\ 
\end{aligned}
\end{equation}
where $\mathbf D\left( {{{\hat \Theta }^{(s)}}} \right)_{wh}$ denotes the output of the domain classifier in each location $(w,h)$. $\alpha$ and $\beta$ are balance coefficients and we set them to 1 following~\cite{Saito_2019_CVPR}.

\noindent{\textbf{Focal Loss (FL).}}
Focal loss $\mathcal{L}_\mathbf{FL}$~\cite{lin2017focal} is adopted to ignore easy-to-classify examples and focus on those hard-to-classify ones during training:
\begin{equation}\label{gamma}
\mathcal{L}_\mathbf{FL}\left(p_{\mathrm{t}}\right)=-f\left(p_{\mathrm{t}}\right) \log \left(p_{\mathrm{t}}\right), f\left(p_{\mathrm{t}}\right)=\left(1-p_{\mathrm{t}}\right)^{\gamma}
\end{equation}
where $p_{\mathrm{t}}=p \text { if } d_i=1, \text{otherwise}, p_{\mathrm{t}}={1-p}$.

\subsection{Ablation Studies from Cityscapes to FoggyCityscapes} \label{ablation}
We first investigate each component and design of our SCL framework from Cityscapes to FoggyCityscapes. Both source and target datasets have  2,975 images in the training set and 500 images in the validation set. We design several controlled experiments for this ablation study. A consistent setting is imposed on all the experiments, unless when some components or structures are examined. In this study, we train models with the ImageNet~\cite{deng2009imagenet} pre-trained ResNet-101 as a backbone, we also provide the results with pre-trained VGG16 model.

The results are summarized in Table~\ref{ablation_foggy}. We present several combinations of four complementary objectives with their loss names and performance. We observe that ``$LS$|$CE$|$FL$|$FL$'' obtains the best accuracy with {\em Context} and {\em Detach}. It indicates that $LS$ can only be placed on the low-level features (rich spatial information and poor semantic information) and $FL$ should be in the high-level locations (weak spatial information and strong semantic information). For the middle location, $CE$ will be a good choice. If you use $LS$ for the middle/high-level features or use $FL$ on the low-level features, it will confuse the network to learn hierarchical semantic outputs, so that {\em ILoss+detach} will lose effectiveness under that circumstance. This verifies that domain adaptive object detection relies heavily on the deep supervision, however, the diverse supervisions should be adopted in a controlled and correct manner. Furthermore, our proposed method performs much better than baseline Strong-Weak~\cite{Saito_2019_CVPR} (37.9\% {\em vs.}34.3\%) and other state-of-the-arts.

\vspace{-0.1cm}
\subsection{Similar Domains}
\vspace{-0.05cm}

\noindent{\textbf{Between Cityspaces and KITTI.}} 
In this part, we focus on studying adaptation between two real and similar domains, as we take KITTI and Cityscapes as our training and testing data. Following~\cite{chen2018domain}, we use KITTI training set which contains 7,481 images. We conduct experiments on both adaptation directions K $\to$ C and C $\to$ K and evaluate our method using AP of {\em car} as in DA.

As shown in Table~\ref{tab:KC}, our proposed method performed much better than the baseline and other state-of-the-art methods. Since Strong-Weak~\cite{Saito_2019_CVPR} didn't provide the results on this dataset, we re-implement it and obtain 37.9\% AP on K$\to$C and 71.0\% AP on C$\to$K. Our method is 4\% higher than the former and 1.7\% higher than latter. If comparing to the non-adapted results (source only), our method outperforms it with a huge margin (about 10\% and 20\% higher, respectively).

\begin{table}[t]
	\caption{Adaptation results between KITTI and Cityscapes. We report AP of {\em Car} on both directions: K$\to$C and C$\to$K. We re-implemented DA~\cite{chen2018domain} and Weak-Strong~\cite{Saito_2019_CVPR} based on the same Faster RCNN framework~\cite{ren2015faster}.}
	\vspace{-.1in}
	\label{tab:KC}
	\centering
	\resizebox{0.25\textwidth}{!}{%
		\begin{tabular}{l|c|c}
			\toprule[1.5pt]
			Method   & K$\to$C & C$\to$K \\ \hline
			Faster RCNN &30.2&53.5\\ \hline
			DA~\cite{chen2018domain}  &38.5&64.1\\ 
			DA (Our impl.)~\cite{chen2018domain}  &35.6&70.8\\ 
			WS (Our impl.)~\cite{Saito_2019_CVPR}  &37.9&71.0\\ \hline
			Ours     &\bf 41.9&\bf 72.7\\
			\bottomrule[1.5pt]
		\end{tabular}
	}
	\vspace{-.05in}
\end{table}

\begin{table}[t]
	\centering
	\caption{Adaptation results on INIT dataset.}
	\vspace{-.2cm}
	\label{tab:INIT}
	\resizebox{0.33\textwidth}{!}{%
		\begin{tabular}{l|l|c|c|c|c}
			\toprule[1.5pt]
			&                            & Car                   &  Sign      & Person                & mAP                   \\ \hline
			\multirow{3}{*}{s2n} & Faster&63.33 &63.96 &32.00 &53.10 \\ 
			& Strong-Weak&67.43&64.33&\bf 32.53&54.76\\ 
			& Ours& \bf 67.92&\bf 65.89&32.52&\bf 55.44\\ \cline{2-6} 
			& Oracle&80.12&84.68&44.57&69.79\\ \hline
			\multirow{3}{*}{s2r} & Faster&70.20 &72.71 &36.22 &59.71 \\ 
			& Strong-Weak&\bf 71.56&78.07&39.27&62.97\\ 
			& ours& 71.41&\bf 78.93&\bf 39.79&\bf 63.37\\ \cline{2-6} 
			& Oracle & 71.83 & 79.42 & 45.21 & 65.49 \\ \hline
			\multirow{3}{*}{s2c}  & Faster&-- &-- &-- &-- \\ 
			&Strong-Weak&\bf 71.32&72.71&43.18&62.40\\ 
			& Ours  & 71.28 &\bf 72.91 &\bf 43.79 & \bf 62.66 \\ \cline{2-6} 
			&Oracle & 76.60 & 76.72 & 47.28 & 66.87\\ 
			\bottomrule[1.5pt]
		\end{tabular}
	}
\end{table}

\begin{table*}[t]
	\caption{Results on adaptation from PASCAL VOC to Clipart Dataset. Average precision (\%) is evaluated on target images.}
	\vspace{-2.8mm}
	\label{tbl:ap_clipart}
	\centering
	\scalebox{0.84}{
		\tabcolsep=1.5pt
		\centering
		\begin{tabular}{l|cccccccccccccccccccc|c}
			\toprule[1.5pt]
			Method &  aero & bcycle & bird & boat & bottle & bus  & car  & cat  & chair & cow  & table & dog  & hrs & bike & prsn & plnt & sheep & sofa & train & tv & \bf mAP  \\\hline
			Faster (Non-adapted)& {35.6}      & 52.5    & 24.3 & 23.0 & 20.0   & 43.9 & 32.8 & 10.7 & 30.6  & 11.7 & 13.8        & 6.0  & \bf{36.8}  & 45.9      & 48.7   & 41.9        &{16.5}  & 7.3  & 22.9  & 32.0      & 27.8 \\
			BDC-Faster &20.2      & 46.4    & 20.4 & 19.3 & 18.7   & 41.3 & 26.5 & 6.4  & 33.2  & 11.7 &{ 26.0}        & 1.7  & 36.6  & 41.5      & 37.7  & 44.5        & 10.6  & 20.4 & 33.3  & 15.5      & 25.6 \\
			DA&15.0&34.6&12.4&11.9&19.8&21.1&23.2&3.1&22.1&26.3&10.6&10.0&19.6&39.4&34.6&29.3&1.0&17.1&19.7&24.8&19.8
			\\\hline
			WST-BSR~\cite{Kim_Self_2019}&28.0&\bf 64.5& 23.9& 19.0& 21.9&\bf 64.3&\bf 43.5& 16.4&\bf 42.2& 25.9&\bf 30.5& 7.9& 25.5& 67.6& 54.5& 36.4& 10.3& \bf 31.2& \bf 57.4& 43.5& 35.7\\
			Strong-Weak~\cite{Saito_2019_CVPR}&26.2&48.5&32.6&\bf{33.7}&38.5&{54.3}& 37.1&{18.6}&34.8&{58.3}&17.0&12.5&33.8&65.5&\bf 61.6&\bf{52.0}&9.3&{24.9}& 54.1&\bf{49.1}&{38.1}\\ \hline
			Ours w/$\mathcal{L}_{ILoss}=FL$ &33.4&49.2&36.0&27.1&38.4&55.7&38.7&15.9&39.0&59.2&18.8&23.7&36.9&70.0&60.6&49.7&25.8&34.8&47.2&51.2&40.6 \\
			Ours w/$\mathcal{L}_{ILoss}=CE$ &\bf 44.7&50.0&33.6&27.4&\bf 42.2&55.6&38.3&\bf 19.2&37.9&\bf 69.0&30.1&\bf 26.3&34.4&67.3&61.0&47.9&21.4&26.3&50.1&47.3&\bf 41.5 \\
			
			\bottomrule[1.5pt]
	\end{tabular}}
	\vspace{-0.15cm}
\end{table*}

\noindent{\textbf{INIT Dataset.}}
INIT Dataset~\cite{shen2019towards} contains 132,201 images for training and 23,328 images for testing. There are four domains: sunny, night, rainy and cloudy, and three instance categories, including: car, person, speed limited sign. This dataset is first proposed for the instance-level image-to-image translation task, here we use it for the domain adaptive object detection purpose.

Our results are shown in Table~\ref{tab:INIT}. Following~\cite{shen2019towards}, we conduct experiments on three domain pairs: sunny$\to$night (s2n), sunny$\to$rainy (s2r) and sunny$\to$cloudy (s2c). Since the training images in rainy domain are much fewer than sunny, for s2r experiment we randomly sample the training data in sunny set with the same number of rainy set and then train the detector. It can be observed that our method is consistently better than the baseline method. We don't provide the results of s2c (faster) because we found that cloudy images are too similar to sunny in this dataset (nearly the same), thus the non-adapted result is very close to the adapted methods.

\subsection{Discrepant Domains}
In this section, we focus on the dissimilar domains, i.e., adaptation from real images to cartoon/artistic. Following~\cite{Saito_2019_CVPR}, we use PASCAL VOC dataset (2007+2012 training and validation combination for training) as the source data and the Clipart or Watercolor~\cite{inoue2018cross} as the target data.
The backbone network is ImageNet pre-trained ResNet-101. 

\noindent{\textbf{PASCAL to Clipart.}} Clipart dataset contains 1,000 images in total, with the same 20 categories as in PASCAL VOC. As shown in Table~\ref{tbl:ap_clipart}, our proposed SCL outperforms all baselines. In addition, we observe that replacing $FL$ with $CE$ loss on instance-context classifier can further improve the performance from 40.6\% to 41.5\%. More ablation results are shown in our Appendix~\ref{app_clipart} (Table~\ref{tbl:appendix_clipart}).

\noindent{\textbf{PASCAL to WaterColor.}}
Watercolor dataset contains 6 categories in common with PASCAL VOC and has totally 2,000 images (1,000 images are used for training and 1,000 test images for evaluation). Results are summarized in Table~\ref{tbl:ap_water}, SCL consistently outperforms other state-of-the-arts.

\subsection{From Synthetic to Real Images}

\noindent{\textbf{Sim10K to Cityscapes.}} Sim 10k dataset~\cite{johnson2016driving} contains 10,000 images for training which are generated by the gaming engine Grand Theft Auto (GTA). Following~\cite{chen2018domain,Saito_2019_CVPR}, we use Cityscapes as target domain and evaluate our models on {\em Car} class. Our result is shown in Table~\ref{tab:Sim10k}, which consistently outperforms the baselines.

\begin{table}[t]
	\centering\small
	\setlength{\tabcolsep}{2pt}
	\caption{Adaptation results from PASCAL to WaterColor.} 
	\vspace{-2mm}
	\label{tbl:ap_water}
	\scalebox{0.9}{
		\tabcolsep=1.5pt
		\begin{tabular}{l|cccccc|c}
			\toprule[1.5pt]
			& \multicolumn{7}{c}{AP on a target domain}\\
			Method & bike & bird & car  & cat  & dog  & prsn &\bf mAP  \\\hline
			Source Only  &68.8 & 46.8 & 37.2 & 32.7 & 21.3 & 60.7   & 44.6 \\ \hline
			BDC-Faster&68.6 & 48.3 & 47.2 & 26.5 & 21.7 & 60.5   & 45.5\\
			DA~\cite{chen2018domain} &75.2 & 40.6 & {48.0} & 31.5 & 20.6 & 60.0   & 46.0\\
			WST-BSR~\cite{Kim_Self_2019} & 75.6 &45.8 & 49.3 & 34.1 & 30.3 & 64.1 & 49.9\\ 
			Strong-Weak~\cite{Saito_2019_CVPR}& \bf{82.3}	& \bf{55.9}&46.5&32.7&{35.5}&\bf{66.7}&{53.3}\\ \hline
			Ours& 82.2&55.1&\bf{51.8}&\bf{39.6}&\bf{38.4}&64.0&\bf{55.2}\\
			\bottomrule[1.5pt]
	\end{tabular}}
\end{table}

\begin{table}[t]
	\caption{Adaptation results on {\em Car} from Sim10k to Cityscapes Dataset (\%). {\em Source Only} indicates the non-adapted results ($\lambda = 0.1$ and $\gamma = 2.0$ are used).}
	\label{tab:Sim10k}
	\vspace{-0.15cm}
	\centering
	\scalebox{0.8}{
		\begin{tabular}{c|c}
			\toprule[1.5pt]
			Method   & AP on Car \\ \hline
			Faster &    34.6       \\ 
			DA~\cite{chen2018domain}       & 38.9     \\ 
			Strong-Weak~\cite{Saito_2019_CVPR}&  40.1         \\ 
			MAF~\cite{he2019multi} &  41.1         \\ \hline
			Ours     & \bf 42.6     \\ \bottomrule[1.5pt]
		\end{tabular}
	}
\end{table}

\section{Analysis}

\noindent{\textbf{Hyper-parameter $\mathbf{K}$.}} Table~\ref{tab:ablation_k} shows the results for sensitivity of hyper-parameter $\mathbf{K}$ in Figure~\ref{framework}. This parameter controls the number of SCL losses and context branches. It can be observed that the proposed method performs best when $\mathbf{K}=3$ on all three datasets.

\begin{table}[h]
	\vspace{-0.2cm}
	\caption{Analysis of hype-parameter $\mathbf{K}$.}
	\label{tab:ablation_k}
	\vspace{-0.25cm}
	\centering
	\resizebox{0.36\textwidth}{!}{%
		\begin{tabular}{l|c|c|c}
			\toprule[1.5pt]
			Method                             & $\mathbf{K}$=2 & $\mathbf{K}$=3 & $\mathbf{K}$=4 \\ \hline
			from Cityscapes to Foggycityscapes & 32.7  & \bf 37.9   &  34.5   \\ \hline
			from PASCAL VOC to Clipart         &  39.0   & \bf 41.5 &  39.3   \\ \hline
			from PASCAL VOC to Watercolor      &  54.7 & \bf 55.2   & 53.4  \\ 
			\bottomrule[1.5pt]
		\end{tabular}
	}
	\vspace{-0.2cm}
\end{table}

\begin{figure}
	\centering
	\includegraphics[width=0.98\columnwidth]{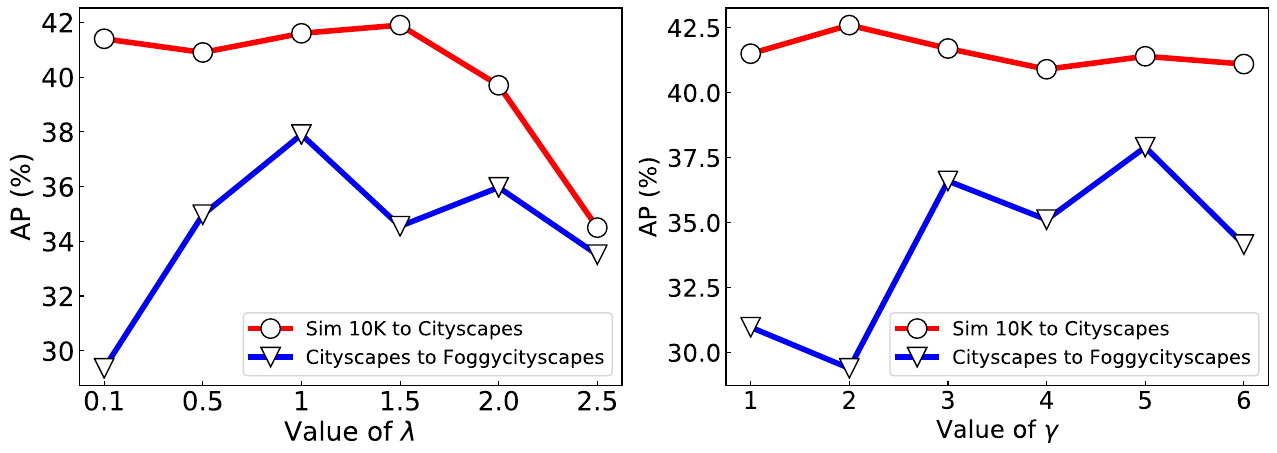}
	\vspace{-0.15cm}
	\caption{Parameter sensitivity for the value of $\lambda$ (left) and $\gamma$ (right) in adaptation from Cityscapes to FoggyCityscapes and from Sim10k to Cityscapes.}
	\label{fig:senti}
	\vspace{-.1cm}
\end{figure}

\begin{figure*}[h]
	\centering
	\subfloat[\scriptsize from Cityscapes and FoggyCityscapes]{\includegraphics[width=0.33\textwidth, keepaspectratio]{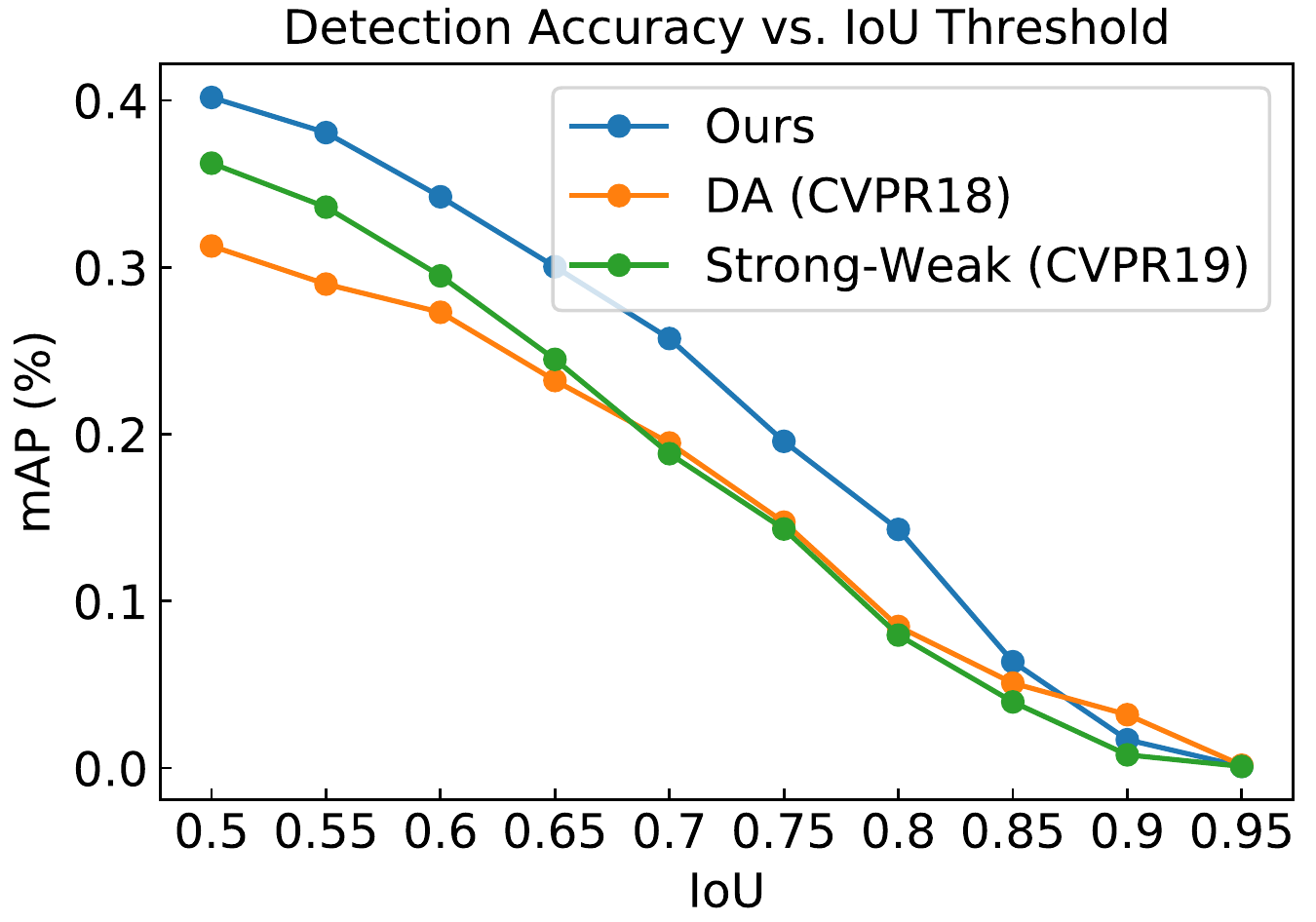}\label{fig:1}}
	\subfloat[\scriptsize from PASCAL VOC to Clipart]{\includegraphics[width=0.33\textwidth, keepaspectratio]{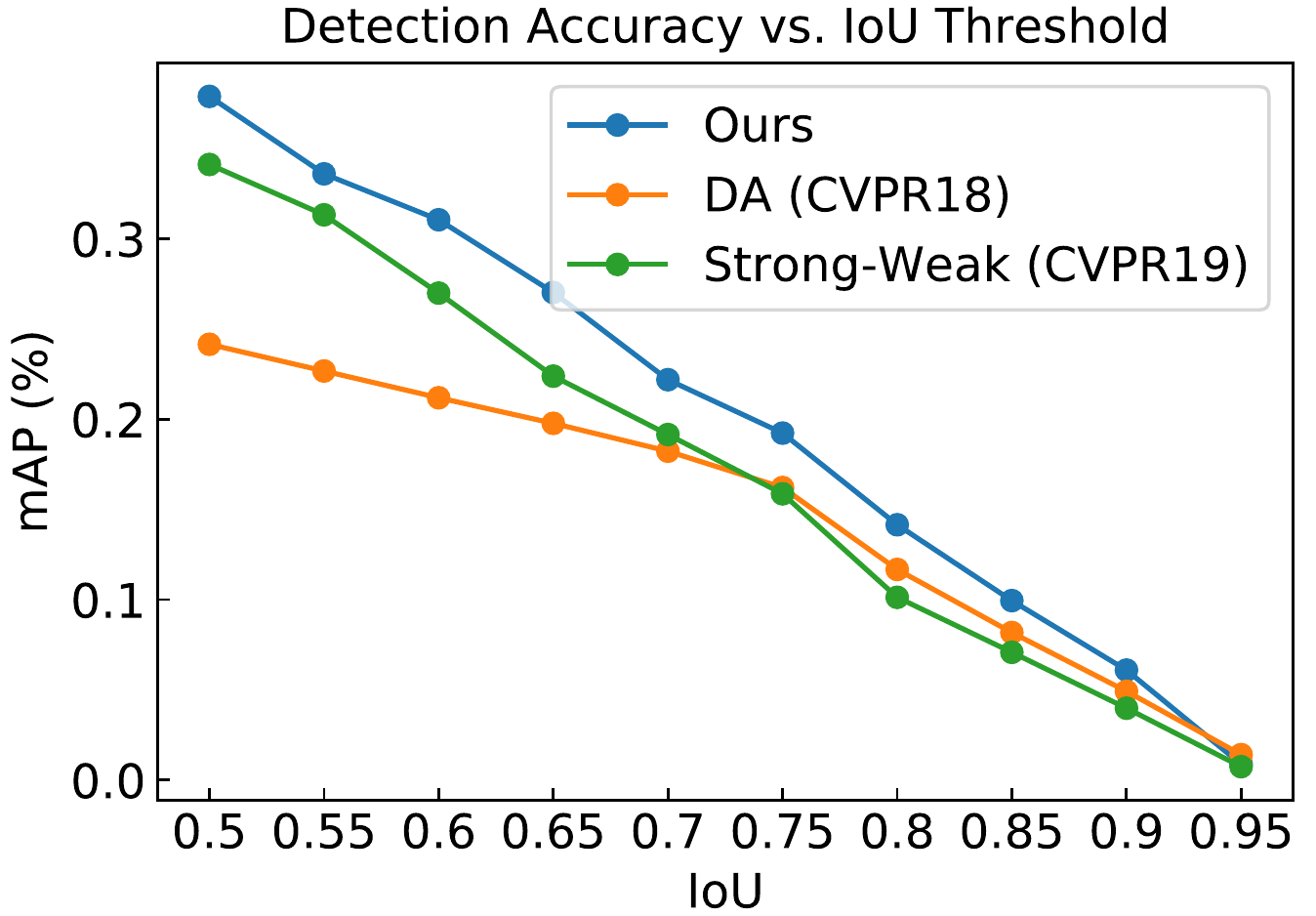}\label{fig:2}}
	\subfloat[\scriptsize from PASCAL VOC to Watercolor]{\includegraphics[width=0.33\textwidth, keepaspectratio]{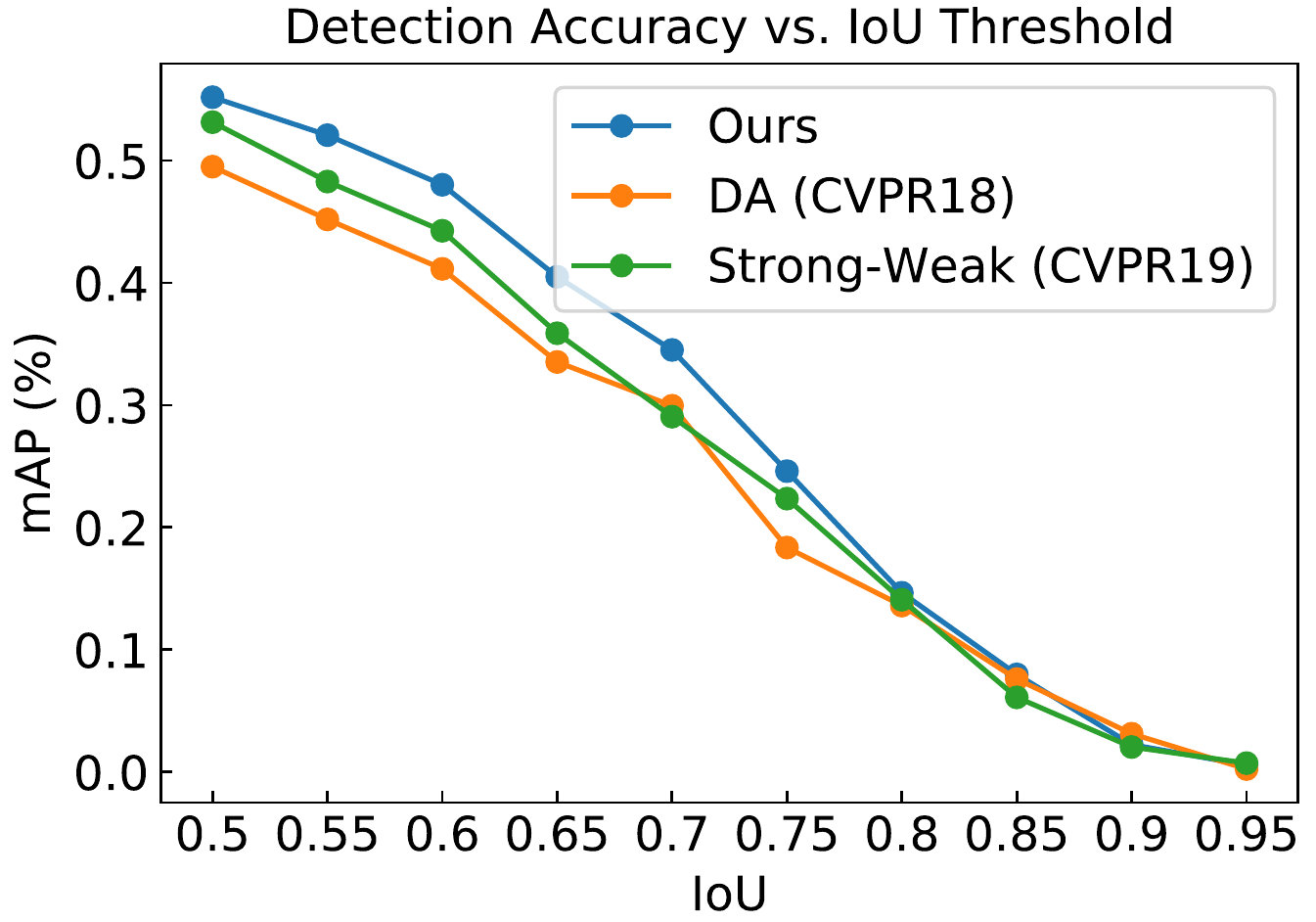}\label{fig:3}}
	\vspace{-0.1in}
	\caption{AP (\%) with different IoU thresholds. We show comparisons on three datasets and all results are calculated with different IoU thresholds and illustrated in different colors.}
	\label{IOU}
	\vspace{-0.1in}
\end{figure*}

\begin{figure*}[t]
	\centering
	\includegraphics[width=0.99\textwidth]{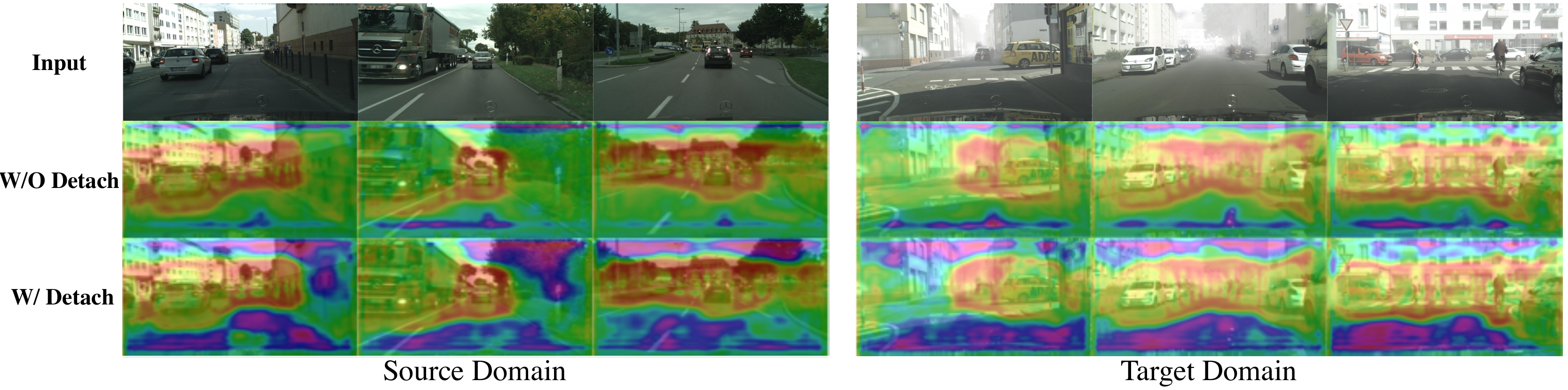}
	\vspace{-0.1in}
	\caption{Visualization of {\em Attention Maps} on source and target domains. We use feature maps after {\bf Conv B3} in Figure~\ref{framework} for visualizing. Top: Input images; Middle: Heatmaps from models {\em w/o} gradient detach; Bottom: Heatmaps from models {\em w/} gradient detach. The colors (red$\to$blue) indicate values from high to low. It can be observed that {\em w/} detach training, our models can learn more discriminative representation between object areas and background (context).}
	\label{heatmaps}
	\vspace{-0.1in}
\end{figure*}

\noindent{\textbf{Parameter Sensitivity on $\lambda$ and $\gamma$.}} Figure~\ref{fig:senti} shows the results for parameter sensitivity of $\lambda$ and $\gamma$ in Eq.~\ref{lambda} and Eq.~\ref{gamma}. $\lambda$ is the trade-off parameter between SCL and detection objectives and $\gamma$ controls the strength of hard samples in {\em Focal Loss}. We conduct experiments on two adaptations: Cityscapes $\to$ FoggyCityscapes (blue) and Sim10K $\to$ Cityscapes (red). On Cityscapes $\to$ FoggyCityscapes, we achieve the best performance when $\lambda=1.0$ and $\gamma=5.0$ and the best accuracy is 37.9\%. On Sim10K $\to$ Cityscapes, the best result is obtained when $\lambda=0.1$, $\gamma=2.0$.

\noindent{\textbf{Analysis of IoU Threshold.}} The IoU threshold is an important indicator to reflect the quality of detection, and a higher threshold means better coverage with ground-truth. In our previous experiments, we use 0.5 as a threshold suggested by many literature~\cite{ren2015faster,chen2018domain}. In order to explore the influence of IoU threshold with performance, we plot the performance {\em vs.} IoU on three datasets. As shown in Figure~\ref{IOU}, our method is consistently better than the baselines on different thresholds by a large margin (in most cases).


\noindent{\textbf{Why Gradient Detach Can Help Our Model?}} To further explore why gradient detach can help to improve performance vastly and what our model really learned, we visualize the heatmaps on both source and target images from our  models {\em w/o} and {\em w/} detach training. As shown in Figure~\ref{heatmaps}, the visualization is plotted with feature maps after {\em Conv B3} in Figure~\ref{framework}. We can observe that the object areas and context from {\em detach}-trained models have stronger contrast than {\em w/o} detach model (red and blue areas). This indicates that {\em detach-based} model can learn more discriminative features from the target object and context. To be more precise, {\em w/o detach} model is attentive more on the background (green color), in contrast, {\em with detach} model is  mainly activated on the object areas with less attention on the background (blue color, i.e., less attention). More visualizations are shown in Appendix~\ref{app_heatmaps} (Figure~\ref{heatmaps_more}).

\begin{figure*}[t]
	\centering
	\subfloat[FoggyCityscapes]{\includegraphics[width=0.96\textwidth, keepaspectratio]{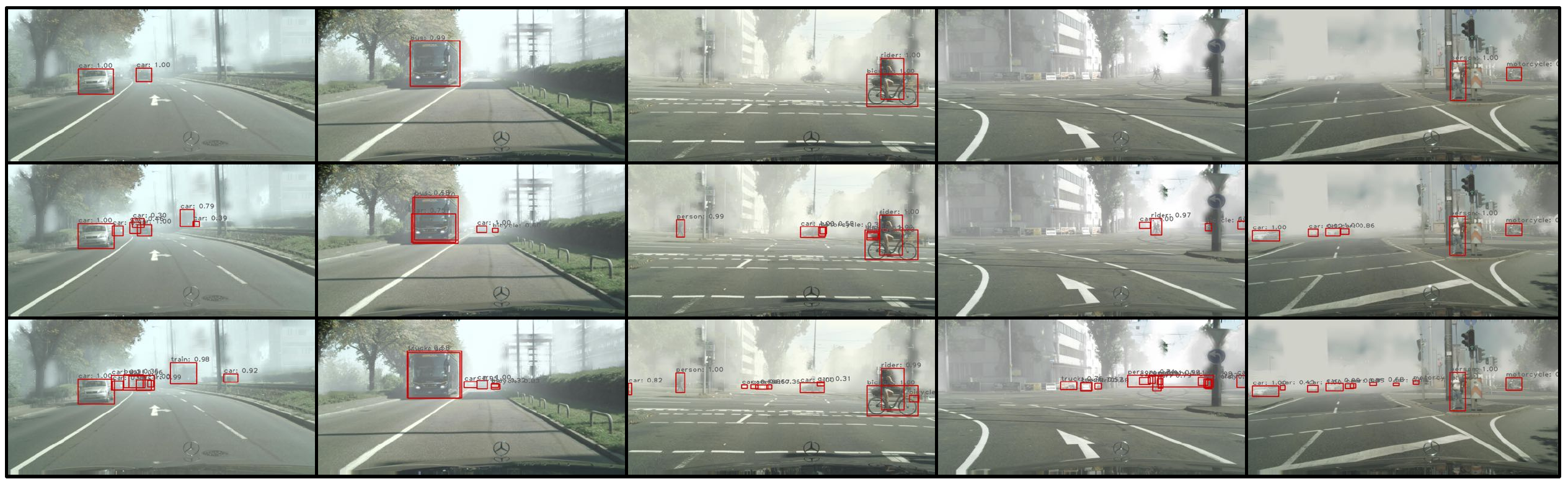}\label{det_fig:1}}\\
	\vspace{-0.12in}
	\subfloat[Clipart]{\includegraphics[width=0.96\textwidth, keepaspectratio]{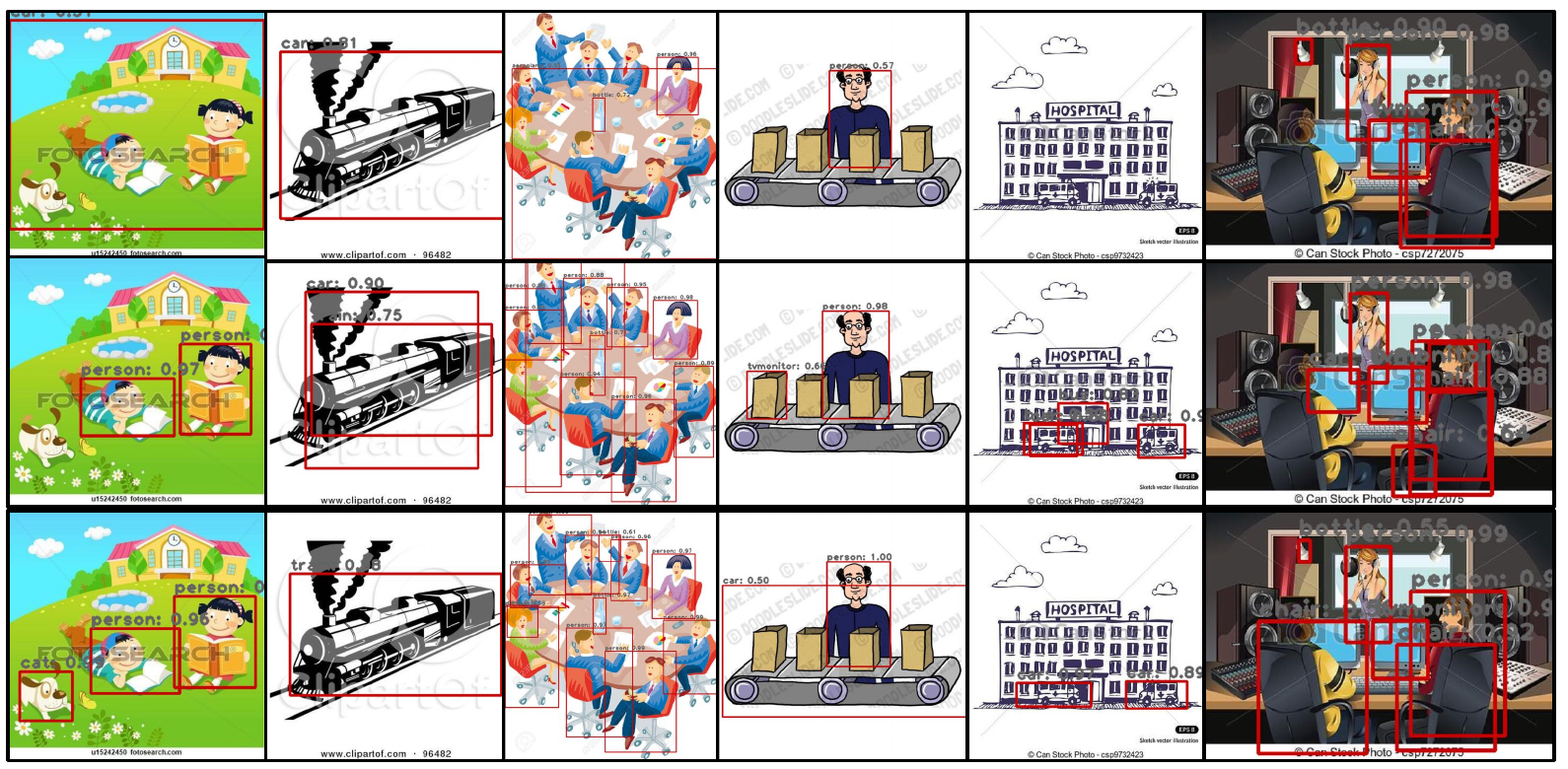}\label{det_fig:2}}\\
	\vspace{-0.12in}
	\subfloat[Watercolor]{\includegraphics[width=0.96\textwidth, keepaspectratio]{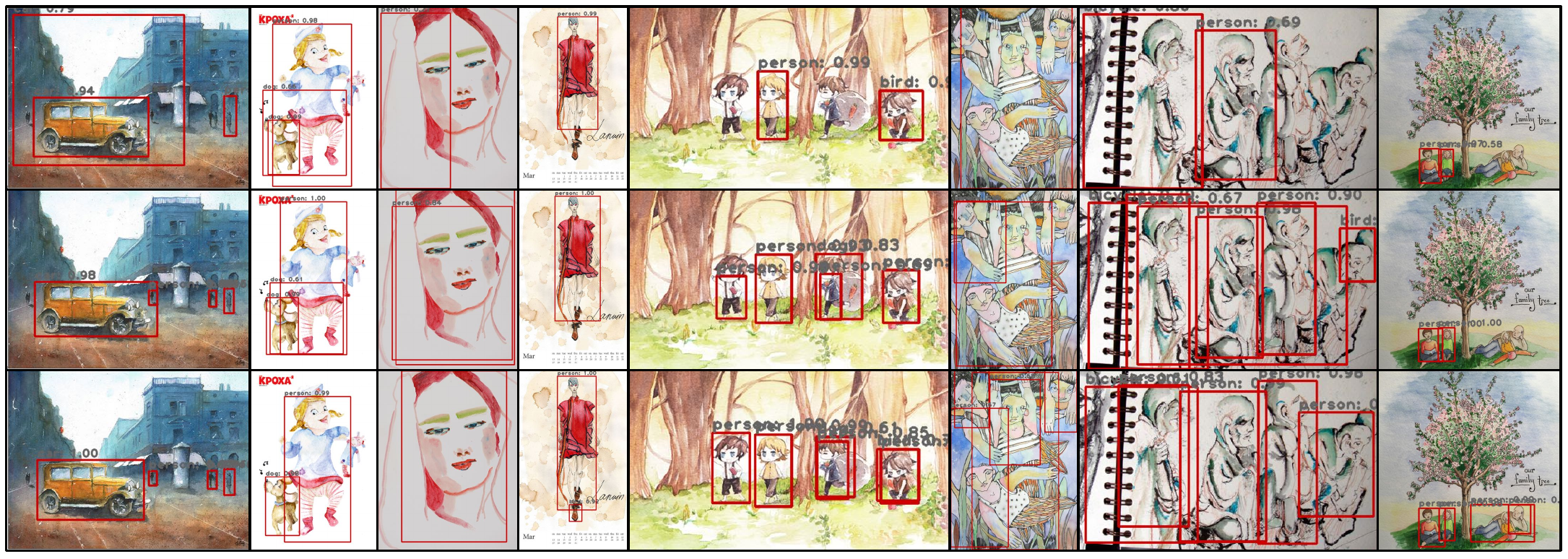}\label{det_fig:3}}
	\vspace{-0.12in}
	\caption{Detection examples with DA~\cite{chen2018domain}, Strong-Weak~\cite{Saito_2019_CVPR} and  our proposed SCL on three datasets. For each group, the first row is the result of DA, the second row is from Strong-Weak and the last row is ours. We show detections with the scores higher than a threshold (0.3 for FoggyCityscapes and 0.5 for other two).}
	\label{detection}
\end{figure*}


\noindent{\textbf{Detection Visualization.}} Figure~\ref{detection} shows several qualitative comparisons of detection examples on three test sets with DA~\cite{chen2018domain}, Strong-Weak~\cite{Saito_2019_CVPR} and our SCL models. Our method detects more small and blurry objects in dense scene (FoggyCityscapes) and suppresses more false positives (Clipart and Watercolor) than the other two baselines.

\section{Conclusion}
We have addressed unsupervised domain adaptive object detection through stacked complementary losses. One of our key contributions is gradient detach training, enabled by suppressing gradients flowing back to the detection backbone. In addition, we proposed to use multiple complementary losses for better optimization. We conduct extensive experiments and ablation studies to verify the effectiveness of each component that we proposed. Our experimental results outperform the state-of-the-art approaches by a large margin on a variety of benchmarks. Our future work will focus on exploring the domain-shift detection from scratch, i.e., without the pre-trained models like DSOD~\cite{shen2017dsod}, to avoid involving bias from the pre-trained dataset.

%

{
\small
\bibliographystyle{ieee_fullname}
\bibliography{my}
}

\clearpage
\begin{appendices}

\section*{\LARGE{Appendix}}

\section{More Ablation Studies}

Table~\ref{tbl:appendix_water} and \ref{tbl:appendix_clipart} show the detailed results on target domains when conducting adaptation from PASCAL VOC to WaterColor and from PASCAL VOC to Clipart dataset. We present results with different combinations of SCL and diverse ablation experiments.

\subsection{From Pascal VOC to Watercolor Dataset}

\begin{table}[h]
\centering
 \caption{AP (\%) on adaptation from PASCAL VOC to WaterColor. ``W/O CLoss ($L_2$)'' means we remove the $L_2$ in our {\em complementary losses}.}  
 \label{tbl:appendix_water}
 \vspace{-0.1in}
 \scalebox{0.87}{
 \tabcolsep=1.5pt
\begin{tabular}{l|cccccc|c}
\toprule[1.5pt]
	 & \multicolumn{7}{c}{AP on a target domain}\\
Method & bike & bird & car  & cat  & dog  & prsn & \bf mAP  \\\hline
$LS$|$CE$|$CE$|$FL$& 76.1&48.8&48.1&29.9&41.2&56.5&50.1\\ 
$LS$|$LS$|$FL$|$FL$& 72.4&51.8&49.7&41.9&36.6&65.5&53.0\\ 
$LS$|$FL$|$FL$|$FL$& 77.8&50.6&48.9&40.1&38.7&63.7&53.3\\ 
$LS$|$CE$|$FL$|$FL$& 82.2&\bf 55.1&\bf {51.8}&{39.6}&{38.4}&64.0&\bf {55.2}\\ 
$LS$|$CE$|$FL$|$CE$& 64.2&54.8&47.3&38.7&\bf 41.7&67.9&52.4 \\ \hline
W/O Detach  &76.2&54.0&49.2&36.7&35.0&\bf 68.6&53.3 \\
W/O ILoss&76.1&51.7&48.0&31.6&40.4&64.3&52.0\\
W/O Context&\bf 83.1&54.5&48.4&34.4&38.8&65.5&54.1\\
W/O Context\&ILoss &69.3&52.8&43.2&\bf 42.7&36.7&66.0&51.8\\
W/O CLoss ($L_2$)& 77.1&53.1&49.6&41.0&39.3&67.9&54.7\\ 
\bottomrule[1.5pt]
\end{tabular}}
\end{table}

\subsection{From Pascal VOC to Clipart Dataset} \label{app_clipart}
The results are shown in Table~\ref{tbl:appendix_clipart}.

\section{Results on Source Domains}
In this section, we show the adaptation results on source domains in Table~\ref{tab:source_detailed-foggy}, \ref{tbl:source_appendix_water}, \ref{tbl:source_appendix_clipart} and \ref{tab:source_KC}. Surprisingly, we observe that the best-trained models (on target domains) are not performing best on the source data, e.g., from PASCAL VOC to WaterColor, DA~\cite{chen2018domain} obtained the highest results on source domain images (although the gaps with Strong-Weak and ours are marginal). We conjecture that the adaptation process for target domains will affect the learning and performing on source domains, even we have used the bounding box ground-truth on source data for training. We will investigate it more thoroughly in our future work and we think the community may also need to rethink whether evaluating on {\em source domain} should be a metric for domain adaptive object detection, since it can help to understand the behavior of models on both source and target images.
\begin{table}[h]
\centering
 \caption{AP ($\%$) of adaptation from Cityscapes to FoggyCityscapes. Results are evaluated on source images (Cityscapes) with the same classes as in the target dataset.} 
\label{tab:source_detailed-foggy}
	\vspace{-0.1in}
 \scalebox{0.68}{
\begin{tabular}{l|c|c|c|c|c|c|c|c|c}
\toprule[1.5pt]
      & \multicolumn{9}{c}{AP on a source domain} \\ \hline
Method  & person & rider & car   & truck & bus   & train & mcycle & bicycle &\bf mAP  \\ \hline
DA (CVPR'18)&33.5&\bf 48.1&51.1&37.0&\bf 61.3&50.0&33.6&36.9&43.9\\ 
Strong-Weak (CVPR'19)&\bf 33.7&47.9&\bf 52.3&33.5&57.1&39.1&35.1&\bf 37.4&42.0\\ 
Ours&33.0&46.7&51.3&\bf 39.8&59.2&\bf 51.6&\bf 36.8&36.5&\bf 44.4 \\ \bottomrule[1.5pt]
\end{tabular}
}
\end{table}

\begin{table}[h]
\centering
 \caption{AP (\%) on adaptation from PASCAL VOC to WaterColor. Results are evaluated on source images (PASCAL VOC) with the same classes as in the WaterColor.} 
 \label{tbl:source_appendix_water}
 	\vspace{-0.1in}
 \scalebox{0.85}{
 \tabcolsep=1.5pt
\begin{tabular}{l|cccccc|c}
\toprule[1.5pt]
	 & \multicolumn{7}{c}{AP on a source domain}\\
Method & bike & bird & car  & cat  & dog  & prsn & \bf mAP  \\\hline
DA (CVPR'18)& \bf 82.0&78.0&86.3&\bf 89.4&\bf 83.5&\bf 82.6&\bf 83.6\\ 
Strong-Weak (CVPR'19)&81.0&77.4&85.3&89.0&82.9&81.4&82.8\\ 
Ours& 80.3&\bf 78.1&\bf 86.5&87.9&\bf 83.5&82.0&83.1\\ 
\bottomrule[1.5pt]
\end{tabular}}

\vspace{0.1in}
\caption{Adaptation results between KITTI and Cityscapes. We report AP of {\em Car} on both directions, including: K$\to$C and C$\to$K of source domain.}
\label{tab:source_KC}
\vspace{-0.1in}
\centering
\scalebox{0.83}{
\begin{tabular}{l|c|c}
\toprule[1.5pt]
Method   & K$\to$C & C$\to$K \\ \hline
DA (CVPR'18)  &\bf 87.9&52.6\\ 
Strong-Weak (CVPR'19) &78.6&\bf 52.9\\ 
Ours     & 78.5& 51.3\\
\bottomrule[1.5pt]
\end{tabular}
}
\end{table}

\section{Detailed Results of Parameter Sensitivity on $\lambda$ and $\gamma$}

We provide the detailed results of parameter sensitivity on $\lambda$ and $\gamma$ in Table~\ref{tab:detailed-lambda} and \ref{tab:detailed-gamma} with the adaptation of from Cityscapes to FoggyCityscapes and from Sim10K to Cityscapes.

\begin{table*}[t]
	
	\caption{AP (\%) on adpatation from PASCAL VOC to Clipart Dataset. Results are evaluated on target images. ``W/O CLoss ($L_2$)'' means we remove the $L_2$ in our {\em complementary losses}.}
	\label{tbl:appendix_clipart}
	\vspace{-0.1in}
	\centering
	\scalebox{0.77}{
		\tabcolsep=1.5pt
		\centering
		\begin{tabular}{l|cccccccccccccccccccc|c}
			\toprule[1.5pt]
			Method &  aero & bcycle & bird & boat & bottle & bus  & car  & cat  & chair & cow  & table & dog  & hrs & bike & prsn & plnt & sheep & sofa & train & tv & \bf mAP  \\\hline
			$LS$|$CE$|$CE$|$FL$ &24.2&48.3&32.6&26.0&31.2&55.3&37.6&12.1&33.0&47.1&23.1&17.0&23.4&57.4&57.3&43.8&19.9&31.7&48.2&42.7&35.4\\
			$LS$|$CE$|$FL$|$CE$ &\bf 44.7&50.0&33.6&27.4&42.2&55.6&38.3&19.2&37.9&\bf 69.0&\bf 30.1&26.3&34.4&67.3&61.0&47.9&21.4&26.3&50.1&47.3&\bf 41.5 \\
			$LS$|$FL$|$FL$|$FL$ &31.4&52.4&31.5&27.5&39.5&56.9&38.4&13.6&38.3&45.5&23.9&15.8&33.7&73.1&\bf 64.6&49.5&19.3&26.8&\bf 55.0&49.9&39.3 \\
			$LS$|$LS$|$FL$|$FL$ &32.3&56.8&33.2&23.8&39.6&46.0&39.6&17.6&38.7&52.4&14.7&21.2&33.0&72.0&59.6&46.7&21.9&26.9&49.2&\bf 51.8&38.9 \\ 
			$LS$|$CE$|$FL$|$FL$ &33.4&49.2&\bf 36.0&27.1&38.4&55.7&38.7&15.9&39.0&59.2&18.8&23.7&36.9&70.0&60.6&\bf 49.7&\bf 25.8&\bf 34.8&47.2&51.2&40.6 \\ \hline
			W/O Detach&33.1&54.5&33.9&\bf 28.2&\bf 45.3&\bf 59.4&31.4&17.4&34.7&39.9&9.8&20.8&33.5&63.0&60.3&40.8&18.7&20.6&51.8&45.6&37.1 \\
			W/O ILoss &27.2&54.0&31.9&24.7&38.6&53.7&36.9&15.1&\bf 40.2&52.4&12.4&\bf 29.6&36.5&69.3&63.6&43.3&20.2&26.9&50.6&44.3&38.6 \\
			W/O Context\&ILoss &38.3&\bf 65.4&25.4&24.6&35.2&47.7&\bf 40.9&\bf 20.9&32.6&29.6&4.6&14.7&26.5&\bf 85.2&60.9&46.6&17.4&22.5&43.9&50.2&36.7\\
			W/O Context&22.5&50.8&33.8&23.5&37.6&48.3&39.4&16.4&38.5&55.7&16.0&23.8&33.0&62.8&59.8&48.4&17.3&28.6&47.6&46.5&37.5\\
			W/O CLoss ($L_2$)&33.1&57.0&32.5&24.6&39.0&55.9&37.3&15.7&39.5&50.7&20.5&19.8&\bf 37.7&75.3&60.8&43.9&21.1&26.2&42.9&45.6&39.0\\ 
			
			\bottomrule[1.5pt]
	\end{tabular}}

		\vspace{0.1in}
	
	\caption{AP (\%) on adaptation from PASCAL VOC to Clipart Dataset. Results are evaluated on source images (PASCAL VOC) with the same classes as in the Clipart.}
	\label{tbl:source_appendix_clipart}
	\vspace{-0.1in}
	\centering
	\scalebox{0.8}{
		\tabcolsep=1.5pt
		\centering
		\begin{tabular}{l|cccccccccccccccccccc|c}
			\toprule[1.5pt]
			Method &  aero & bcycle & bird & boat & bottle & bus  & car  & cat  & chair & cow  & table & dog  & hrs & bike & prsn & plnt & sheep & sofa & train & tv & \bf mAP  \\\hline
			DA &\bf 79.7&\bf 83.2&\bf 81.3&\bf 70.0&\bf 66.6&86.0&\bf 87.3&87.1&57.3&\bf 85.3&\bf 68.5&\bf 87.0&\bf 86.6&\bf 82.3&\bf 80.7&49.7&\bf 80.5&\bf 75.5&82.9&\bf 81.6&\bf 78.0\\
			Strong-Weak&74.3&78.6&66.4&52.7&54.5&80.1&81.4&77.6&43.1&72.9&65.1&74.6&76.5&77.0&75.2&46.3&71.6&64.1&77.0&70.1&69.0\\
			Ours&78.4&81.7&78.4&69.4&60.8&\bf 86.4&86.0&\bf 87.7&\bf 57.9&84.8&68.2&86.4&84.6&82.2&79.3&\bf 50.5&79.9&73.8&\bf 84.2&75.2&76.8\\ 
			\bottomrule[1.5pt]
	\end{tabular}}

\end{table*}

\vspace{-1mm}
\begin{table*}[t]
	\caption{Architectures of the forward networks.}
	\label{forward}
	\vspace{-0.1in}
	\begin{minipage}[t]{.33\textwidth}
		\begin{center}
			\scalebox{0.80}{
				\begin{tabular}{|c|}
					\hline
					\multicolumn{1}{|c|}{Forward Net1} \\
					\hline
					Conv $3 \times 3 \times 256$, stride 1, pad 1\\
					ReLU\\
					Conv $3 \times 3 \times 128$, stride 1, pad 1\\ 
					ReLU\\
					Conv $3 \times 3 \times 128$, stride 1, pad 1\\
					ReLU\\\hline
				\end{tabular}
			}
		\end{center}
	\end{minipage}
	\begin{minipage}[t]{.33\textwidth}
		\begin{center}
			\scalebox{0.82}{
				\begin{tabular}{|c|}
					\hline
					\multicolumn{1}{|c|}{Forward Net2} \\
					\hline
					Conv $3 \times 3 \times 256$, stride 1, pad 1\\
					ReLU\\
					Conv $3 \times 3 \times 128$, stride 1, pad 1\\  
					ReLU\\
					Conv $3 \times 3 \times 128$, stride 1, pad 1\\  
					ReLU\\\hline
				\end{tabular}
			}
		\end{center}
	\end{minipage}
	\begin{minipage}[t]{.33\textwidth}
		\begin{center}
			\scalebox{0.82}{
				\begin{tabular}{|c|}
					\hline
					\multicolumn{1}{|c|}{Forward Net3} \\
					\hline
					Conv $3 \times 3 \times 512$, stride 1, pad 1\\
					ReLU\\
					Conv $3 \times 3 \times 128$, stride 1, pad 1\\  
					ReLU\\
					Conv $3 \times 3 \times 128$, stride 1, pad 1\\  
					ReLU\\\hline
				\end{tabular}
			}
		\end{center}
	\end{minipage}
\end{table*}

\begin{table}[t]
\centering
 \caption{AP ($\%$) of adaptation from Cityscapes to FoggyCityscapes with different $\lambda$ and $\gamma$.} 
\label{tab:detailed-lambda}
\vspace{-0.1in}
	\scalebox{0.82}{
\begin{tabular}{c|cccccccc|c}
\toprule[1.5pt]
      & \multicolumn{9}{c}{AP on a target domain} \\ \hline
$\lambda$ & person & rider & car   & truck & bus   & train & mcycle & bicycle &\bf mAP  \\ \hline
\bf 0.1&25.8&37.2&24.6&24.2&42.0&33.6&17.5&29.9& 29.4  \\ 
\bf 0.5&29.5&42.2&44.4&24.4&45.3&34.1&27.2&32.8&35.0 \\ 
\bf 1.0&30.7&44.1&44.3&30.0&47.9&42.9&29.6&33.7& \bf 37.9\\ 
\bf 1.5&26.3&42.2&43.6&25.5&43.8&36.4&26.7&32.0&34.6  \\ 
\bf 2.0&29.5&39.4&43.7&28.7&46.0&39.7&28.7&32.0&36.0  \\ 
\bf 2.5&25.9&40.3&43.3&26.1&40.8&35.2&26.2&30.2& 33.5 \\ \hline
$\gamma$ & \multicolumn{9}{c}{}                    \\ \hline
\bf 1&27.1&41.6&41.3&25.5&41.6&20.3&20.5&30.0& 31.0\\ 
\bf 2&27.8&41.3&36.4&24.2&38.8&12.8&22.9&30.9& 29.4 \\ 
\bf 3&29.8&40.7&43.9&29.0&45.0&41.5&30.8&32.0&36.6 \\ 
\bf 4&30.3&42.6&44.2&25.4&45.7&33.9&28.6&30.3&35.1\\ 
\bf 5&30.7&44.1&44.3&30.0&47.9&42.9&29.6&33.7&\bf 37.9 \\ 
\bf 6&26.4&42.0&43.8&23.6&45.2&35.2&26.7&30.3&34.2 \\ \bottomrule[1.5pt]
\end{tabular}
}
\end{table}

\begin{table}[h]
\centering
\caption{AP ($\%$) of adaptation from Sim10K to Cityscapes with different $\lambda$ and $\gamma$.}
\vspace{-0.1in}
\label{tab:detailed-gamma}
\scalebox{0.85}{
\begin{tabular}{c|c|c|c|c|c}
\toprule[1.5pt]
\multicolumn{6}{c}{AP on a target domain} \\ \hline
\multicolumn{6}{c}{$\lambda$}                \\ \hline
\bf 0.1   &\bf 0.5   &\bf 1.0   &\bf 1.5   &\bf 2.0  &\bf 2.5  \\ \hline
41.4&40.9&41.6&\bf 41.9&39.7&34.5  \\ \hline
\multicolumn{6}{c}{$\gamma$}                 \\ \hline
\bf 1     &\bf  2     &\bf 3     &\bf 4     &\bf 5    &\bf 6    \\ \hline
41.5&\bf 42.6&41.7&40.9&41.4&41.1  \\ \bottomrule[1.5pt]
\end{tabular}
}
\end{table}

\section{Context Network}
\label{sec:sfp}

Our context networks are shown in Table~\ref{forward}. We use three branches (forward networks) to deliver the context information and each branch generates a 128-dimension feature vector from the corresponding backbone layers of SCL. Then we naively concatenate them and obtain the final context feature with a 384-dimension vector.


\section{Visualization of Intermediate Feature Embedding}
\begin{figure*}[b]
	\centering
	\subfloat[from Cityscapes to FoggyCityscapes]{\includegraphics[width=1\textwidth, keepaspectratio]{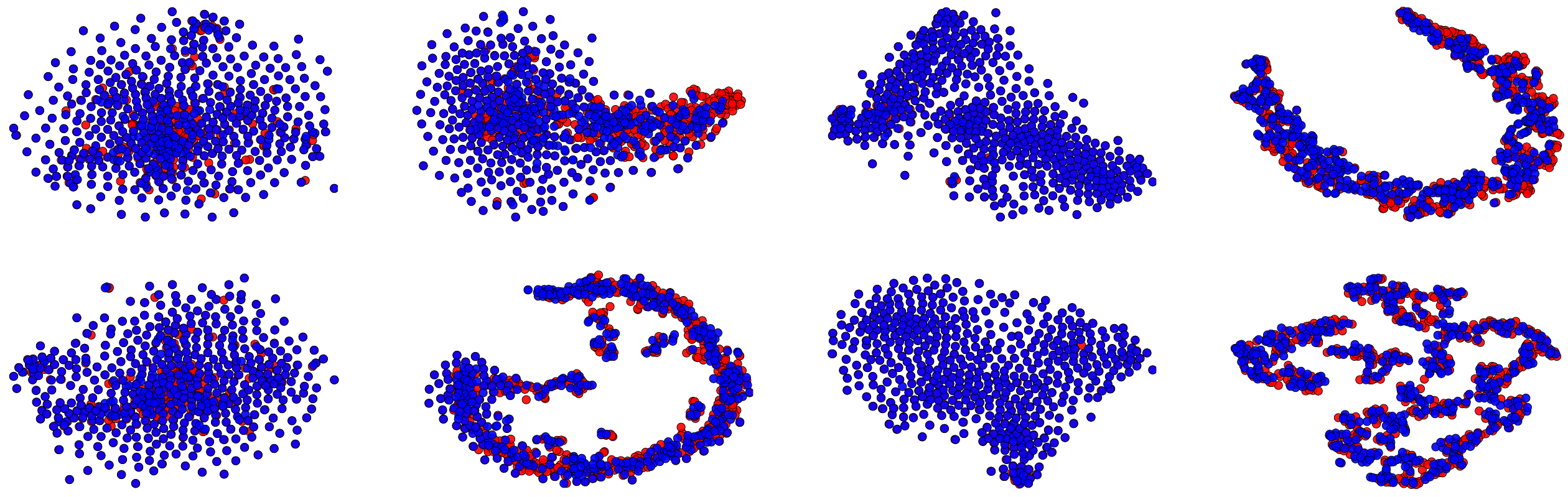}\label{e_fig:1}}\\
	\vspace{-0.1in}
	\subfloat[from PASCAL to Watercolor]{\includegraphics[width=1\textwidth, keepaspectratio]{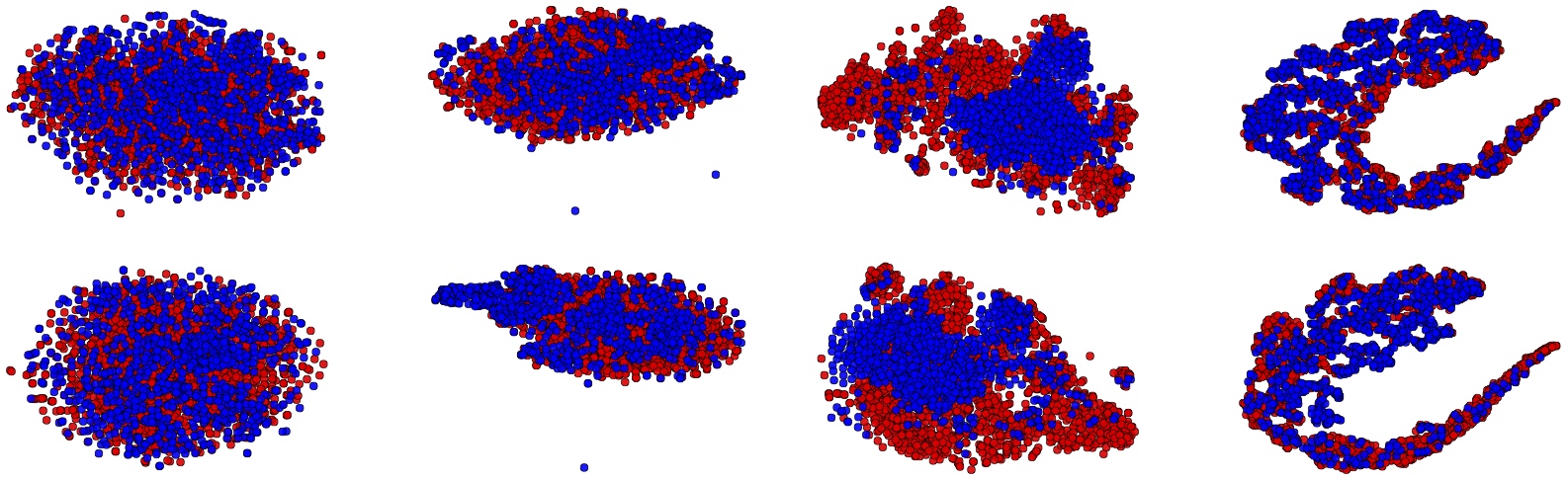}\label{e_fig:2}}\\
	\vspace{-0.1in}
	\subfloat[from PASCAL to Clipart]{\includegraphics[width=1\textwidth, keepaspectratio]{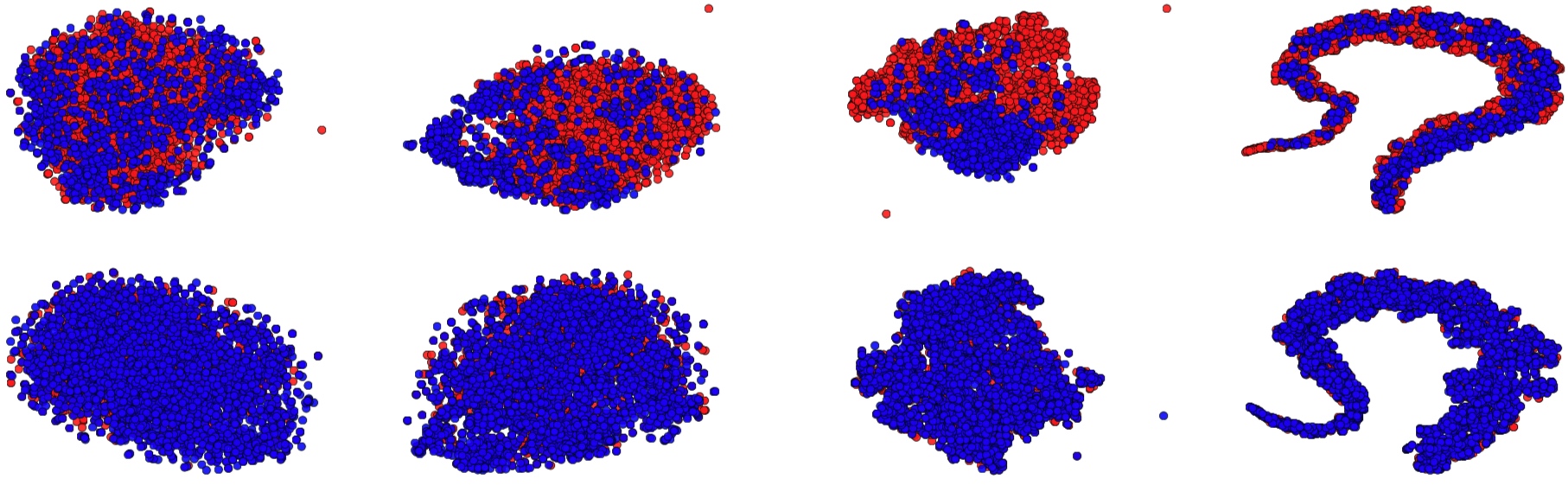}\label{e_fig:3}}
	\vspace{-0.1in}
	\caption{Visualization of feature embedding on three adaptation datasets by t-SNE~\cite{maaten2008visualizing}.
		Red indicates the source examples and blue indicates the target one. In each group, the first row is the result of {\em w/o detach} model, the second row is from {\em with detach} model. In each row, from left to right are results from features after $\text B_1$, $\text B_2$, $\text B_3$ and the 384-dim context features.}
	\label{embedding_appendix}
\end{figure*}

In this section, we visualize the intermediate feature embedding on three adaptation datasets. As shown in Figure~\ref{embedding_appendix}, the gradient {\em detach}-based models can adapt source and target images to a similar distribution better than {\em w/o detach} models. 

\begin{figure*}[h]
	\centering
	\includegraphics[width=1.0\textwidth]{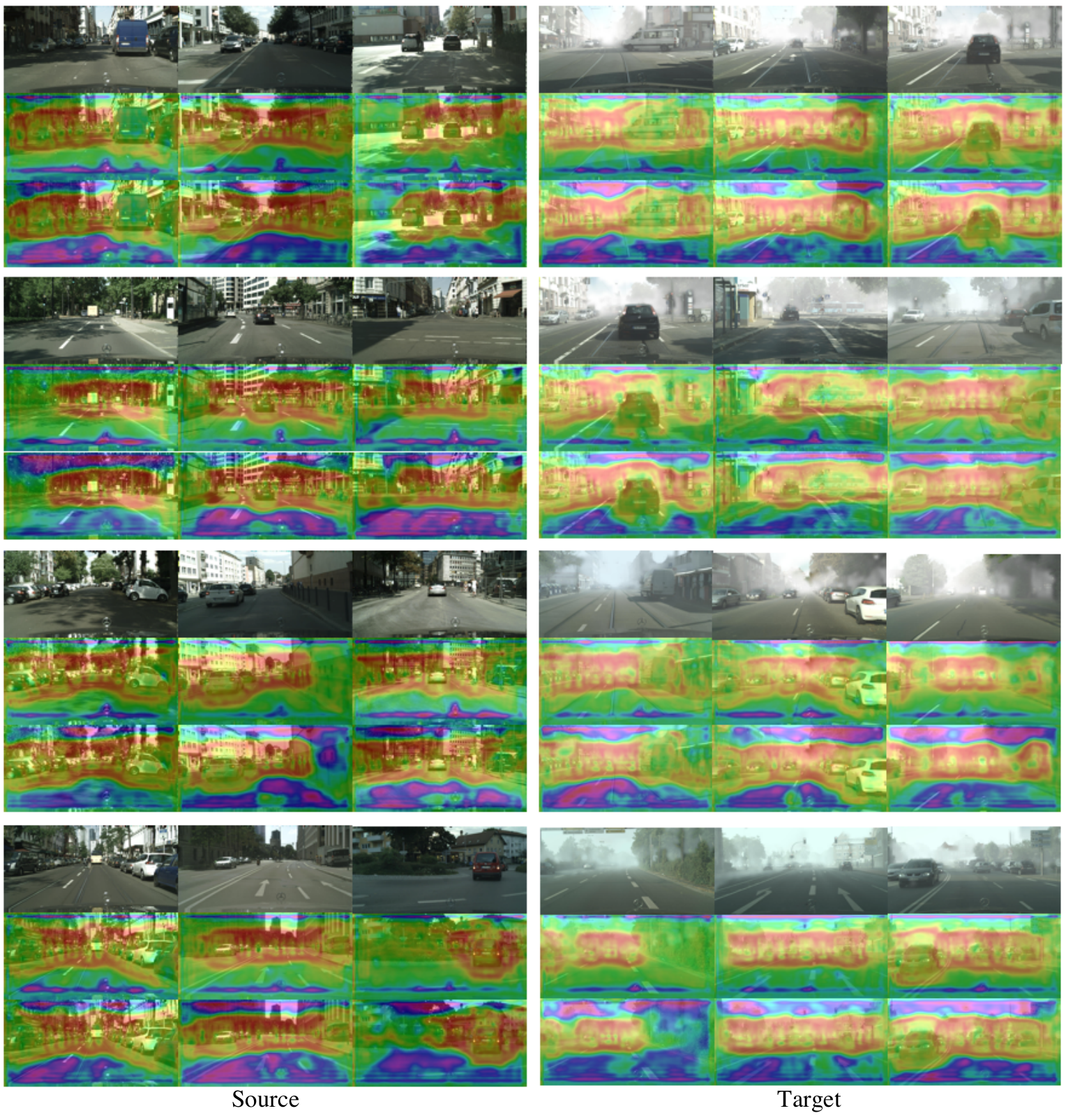}
	\vspace{-0.2in}
	\caption{More visualizations of {\em Attention Maps} on source and target domains. Top: Input images; Middle: Heatmaps from models {\em w/o gradient detach}; Bottom:  Heatmaps from models {\em w/ gradient detach}. The colors (red$\to$blue) indicate values from high to low.}
	\label{heatmaps_more}
	\vspace{-0.2in}
\end{figure*}

\section{More Visualizations of Heatmaps in Fig.~\ref{heatmaps_more}}  \label{app_heatmaps}

\section{More Detection Visualization in Fig.~\ref{detection}}
\begin{figure*}[h]
\centering
\vspace{-0.1in}
\subfloat[Clipart]{\includegraphics[width=1\textwidth, keepaspectratio]{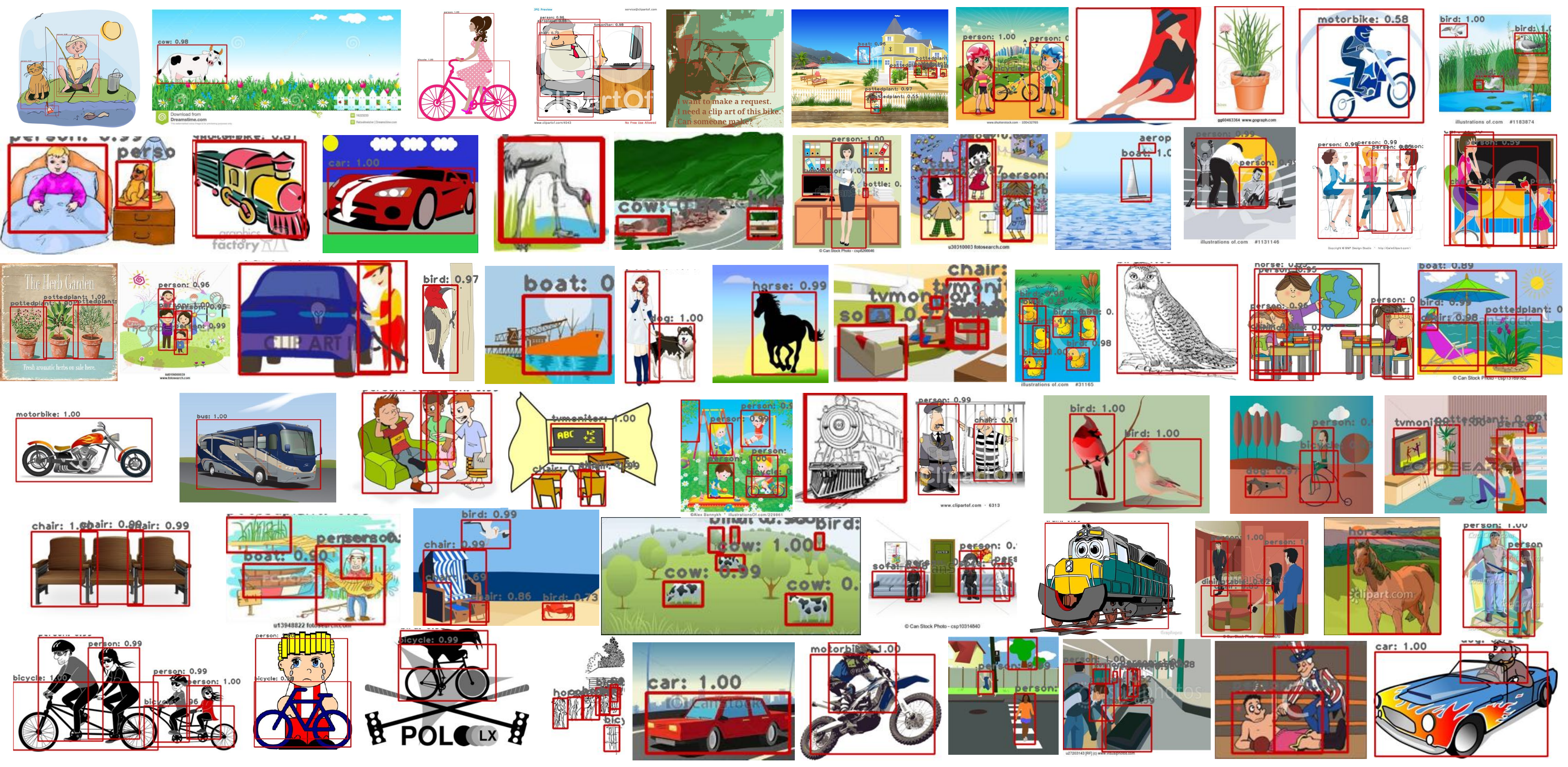}\label{det_fig:2}}\\
\vspace{-0.1in}
\subfloat[Watercolor]{\includegraphics[width=1\textwidth, keepaspectratio]{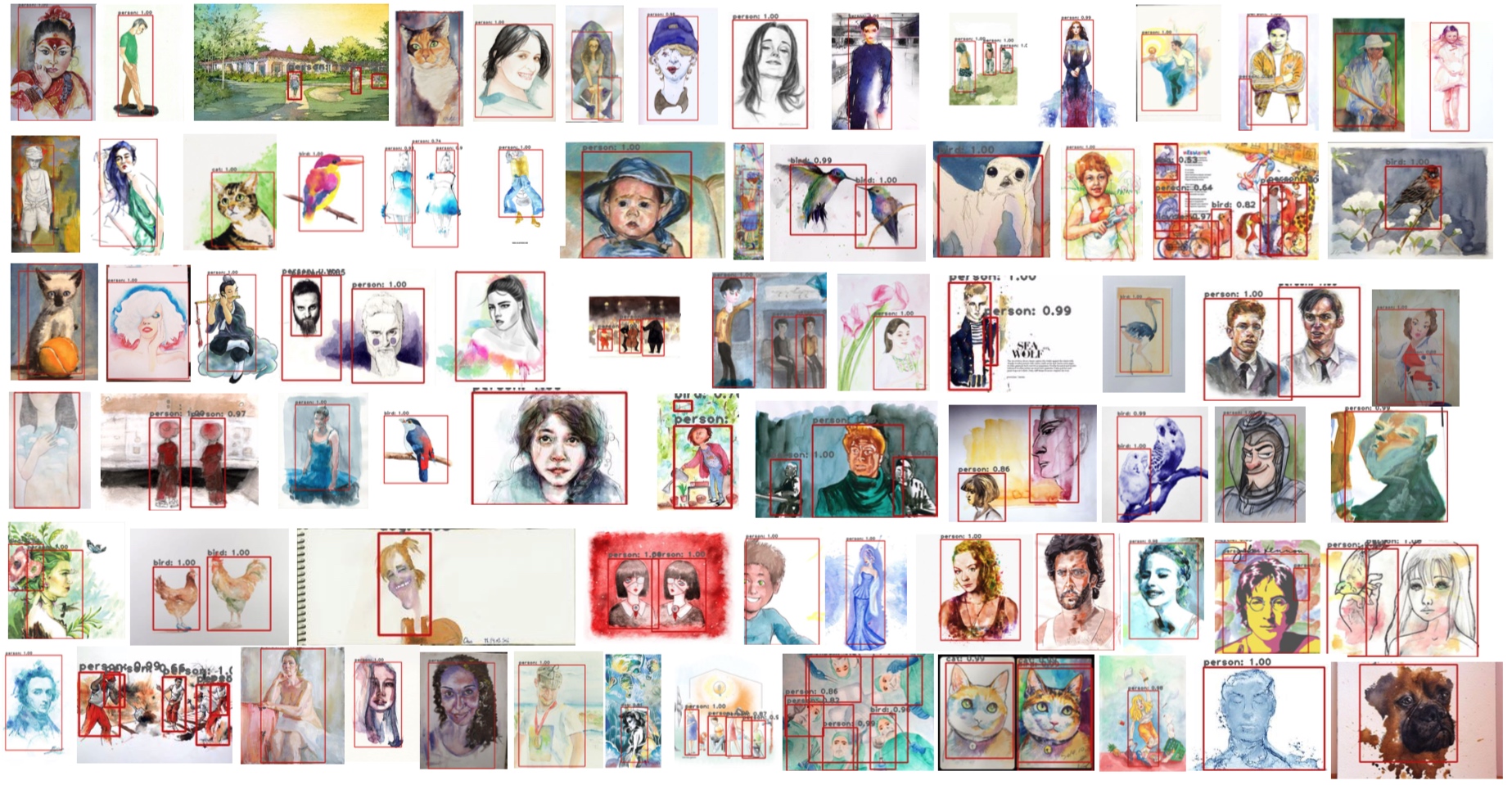}\label{det_fig:3}}
\vspace{-0.1in}
\caption{More detection examples with our proposed SCL on Clipart and Watercolor. We show detections with the scores higher than 0.5.}
\label{detection}
\end{figure*}

\end{appendices}

\end{document}